\documentclass[11pt]{article}

\usepackage[]{emnlp2021}

\usepackage{times}
\usepackage{latexsym}

\usepackage[T1]{fontenc}

\usepackage[utf8]{inputenc}

\usepackage{microtype}

\usepackage[absolute,overlay]{textpos}

\usepackage{multirow}
\usepackage{arydshln}
\usepackage{graphicx}
\usepackage{booktabs}
\usepackage{amssymb}  
\usepackage{subcaption}
\usepackage{amsmath} 

\usepackage{diagbox}
\usepackage{subcaption}

\usepackage{pifont}

\usepackage{xspace}
\usepackage{arydshln}

\newcommand{\cmark}{\ding{51}}%
\newcommand{\xmark}{\ding{55}}%

\newcommand{\EmbLex}[1]{\textsc{EmbLex}}
\newcommand{\EmbNoLex}[1]{\textsc{EmbNoLex}}
\newcommand{\AdapNoMFNoLex}[1]{\textsc{AdapNoMFNoLex}}
\newcommand{\AdapMFNoLex}[1]{\textsc{AdapMFNoLex}}
\newcommand{\AdapNoMFLex}[1]{\textsc{AdapNoMFLex}}
\newcommand{\AdapMFLex}[1]{\textsc{AdapMFLex}}

\newcommand{\elrand}{\textsc{EL-rand}\xspace}
\newcommand{\ellex}{\textsc{EL-lex}\xspace}
\newcommand{\mfrand}{\textsc{MF$^1$-rand}\xspace}
\newcommand{\mflex}{\textsc{MF$^1$-lex}\xspace}

\newcommand{\mfrandkmeans}{\textsc{MF$^{10}_{\textsc{KMeans}}$-rand}\xspace}
\newcommand{\mflexkmeans}{\textsc{MF$^{10}_{\textsc{KMeans}}$-lex}\xspace}

\newcommand{\mfrandnn}{\textsc{MF$^{10}_{\textsc{Neural}}$-rand}\xspace}
\newcommand{\mflexnn}{\textsc{MF$^{10}_{\textsc{Neural}}$-lex}\xspace}

\newcommand{\madx}{\textsc{MAD-X}\xspace}
\newcommand{\madxtwo}{\textsc{MAD-X 2.0}\xspace}

\newcommand{\unk}{\texttt{UNK}\xspace}

\DeclareMathOperator*{\argmax}{arg\,max}

\title{UNKs Everywhere: \\ Adapting Multilingual Language Models to New Scripts}

\author{Jonas Pfeiffer$^{1}$, Ivan Vuli\'{c}$^{2}$, {\bf Iryna Gurevych$^{1}$, Sebastian Ruder$^{3}$ } \\
$^1$Ubiquitous Knowledge Processing Lab, 
  Technical University of Darmstadt \\
$^2$Language Technology Lab, University of Cambridge \hspace{0.5em} \\
$^3$DeepMind \\
\texttt{pfeiffer@ukp.tu-darmstadt.de} \\
}

\date{}

\begin{document}
\maketitle
\begin{abstract}
Massively multilingual language models such as multilingual BERT offer state-of-the-art cross-lingual transfer performance on a range of NLP tasks. However, due to limited capacity and large differences in pretraining data sizes, there is a profound performance gap between resource-rich and resource-poor target languages. 
The ultimate challenge is dealing with under-resourced languages not covered at all by the models and written in scripts \textit{unseen} during pretraining. In this work, we propose a series of novel data-efficient methods that enable quick and effective adaptation of pretrained multilingual models to such low-resource languages and unseen scripts. Relying on matrix factorization, our methods capitalize on the existing latent knowledge about multiple languages already available in the pretrained model's embedding matrix. Furthermore, we show that learning of the new dedicated embedding matrix in the target language can be improved by leveraging a small number of vocabulary items (i.e., the so-called \textit{lexically overlapping} tokens) shared between mBERT's and target language vocabulary. Our adaptation techniques offer substantial performance gains for languages with unseen scripts. We also demonstrate that they can yield improvements for low-resource languages written in scripts covered by the pretrained model.
\end{abstract}

\section{Introduction}
\begin{figure}[!t]
    \centering
        \centering
        \includegraphics[width=0.7\linewidth]{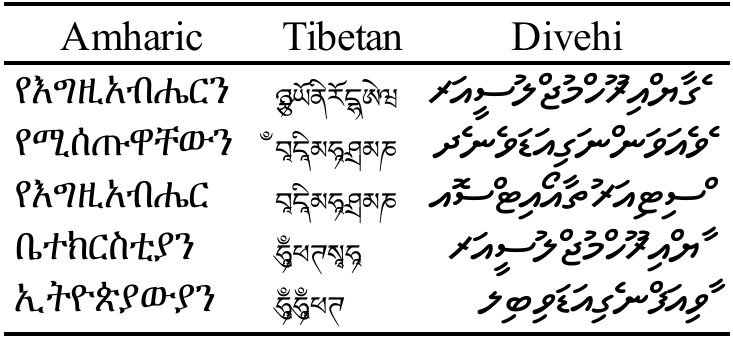}
    \vspace{-1mm}
    \caption{Example tokens of unseen scripts.}
    \vspace{-1.5mm}
\label{fig:script_examples}
\vspace{-1.0em}
\end{figure}

Massively multilingual language models pretrained on large multilingual data, such as multilingual BERT \cite[mBERT;][]{Devlin2019bert} and XLM-R \cite{Conneau2020xlm-r} are the current state-of-the-art vehicle for effective cross-lingual transfer \cite{Hu2020xtreme}. However, while they exhibit strong transfer performance between resource-rich and similar languages \cite{Conneau2020xlm-r,Artetxe2020cross-lingual}, these models struggle with transfer to low-resource languages \cite{Wu:2020repl} and languages not represented at all in their pretraining corpora \cite{pfeiffer20madx,Muller20BeingUnseen,Ansell2021MADG}. The most extreme challenge is dealing with \textit{unseen languages with unseen scripts} (i.e., the scripts are not represented in the pretraining data; see Figure \ref{fig:script_examples}), where the pretrained models are bound to fail entirely if they are used off-the-shelf without any \textit{further model adaptation}.

Existing work focuses on the embedding layer and learns either a new embedding matrix for the target language \cite{Artetxe2020cross-lingual} or adds new tokens to the pretrained vocabulary. While the former has only been applied to high-resource languages, the latter approaches have been limited to languages with seen scripts \cite{ChauLS20Parsing, Muller20BeingUnseen} and large pretraining corpora \cite{Wang20ExtendmBERT}. Another line of work adapts the embedding layer as well as other layers of the model via adapters \cite{pfeiffer20madx, ustun2020udapter}. Such methods, however, cannot be directly applied to languages with unseen scripts.

In this work, we first empirically verify that the original tokenizer and the original embedding layer of a pretrained multilingual model fail for languages with unseen script. This implies that dedicated in-language tokenizers and embeddings are a crucial requirement for any successful model adaptation. The key challenge is aligning new target language embeddings to the pretrained model's representations while leveraging knowledge encoded in the existing embedding matrix.
We systematize existing approaches based on the pretrained information they utilize and identify \emph{lexically overlapping tokens} that are present in both vocabularies as key carriers of such information \cite{Sogaard2018limitations}.\footnote{Even languages with unseen scripts share some tokens, e.g. numbers, foreign named entities written in their original scripts, etc. with seen languages.}
We then present novel, effective, and data-efficient methods for adapting pretrained multilingual language models to resource-low languages written in different scripts. Beyond lexical overlap, our methods rely on factorized information from the embedding matrix and token groupings. 

We evaluate our approaches in the named entity recognition (NER) task on the standard WikiAnn dataset \cite{Rahimi2019massively} and Dependency Parsing \cite[DP;][]{nivre:2016}. 
We use 4 diverse resource-rich languages as source languages, and transfer to 17 and 6 resource-poor target languages respectively, including 5 languages with unseen scripts (Amharic, Tibetan, Khmer, Divehi, Sinhala).
We show that our adaptation techniques offer unmatched performance for languages with unseen scripts. They also yield improvements for low-resource and under-represented languages written in scripts covered by the pretrained model. 

\vspace{1.4mm}
\noindent \textbf{Contributions.} \textbf{1)}~We systematize and compare current model adaptation strategies for low-resource languages with seen and unseen scripts. \textbf{2)}~We measure the impact of initialization when learning new embedding layers, and demonstrate that non-random initialization starting from a subset of seen lexical items (i.e., \textit{lexically overlapping} vocabulary items) has a strong positive impact on task performance for resource-poor languages. \textbf{3)}~We propose methods for learning low-dimensional embeddings, which reduce the number of trainable parameters and yield more efficient model adaptation. Our approach, based on matrix factorization and language clusters, extracts relevant information from the pretrained embedding matrix. \textbf{4)}~We show that our methods outperform previous approaches  with both resource-rich and resource-poor languages. They substantially reduce the gap between random and lexically-overlapping initialization, enabling better model adaption to unseen scripts.

The code for this work is released at
\href{https://github.com/Adapter-Hub/UNKs_everywhere}{github.com/ Adapter-Hub/UNKs\_everywhere}.

\vspace{-0.3em}
\section{Background: Multilingual Model Adaptation for Cross-lingual Transfer}
\label{s:adaptation}
\label{s:rw}
\vspace{-0.3em}

Recent language models \cite{Devlin2019bert, Conneau2020xlm-r}, based on Transformer architectures \cite{Vaswani2017transformer} and pretrained on massive amounts of multilingual data, have recently surpassed (static) cross-lingual word embedding spaces \cite{Ruder:2019jair,Glavas:2019acl} as the state-of-the-art paradigm for cross-lingual transfer in NLP \cite{Pires2019,Wu2019beto,Wu2020emerging,Hu2020xtreme,K2020Crosslingab}.  However, recent studies have also indicated that even current state-of-the-art models such as XLM-R (Large) still do not yield reasonable transfer performance across a large number of target languages \cite{Hu2020xtreme}. The largest drops are reported for resource-poor target languages \cite{Lauscher:2020zerohero}, and (even more dramatically) for languages not covered at all during pretraining \cite{pfeiffer20madx}. 

\vspace{1mm}
\noindent \textbf{Standard Cross-Lingual Transfer Setup} 
with a state-of-the-art pretrained multilingual model such as mBERT or XLM-R is \textbf{1)} fine-tuning it on labelled data of a downstream task in a source language and then \textbf{2)} applying it directly to perform inference in a target language \cite{Hu2020xtreme}. However, as the model must balance between many languages in its representation space, it is not suited to excel at a specific language at inference time without further adaptation \cite{pfeiffer20madx}.

\vspace{1mm}
\noindent \textbf{Adapters for Cross-lingual Transfer.}
Adapter-based approaches have been proposed as a remedy \cite{Rebuffi2017adapters,Rebuffi2018, Houlsby2019adapters,Cooper2019adapters, Bapna2019adapters,  pfeiffer2020AdapterHub, Pfeiffer2020adapterfusion}. In the cross-lingual setups, the idea is to increase the multilingual model capacity by storing language-specific knowledge of each language in dedicated parameters \cite{pfeiffer20madx, Vidoni2020OrthogonalLA}. We start from MAD-X \cite{pfeiffer20madx}, a state-of-the-art adapter-based framework for cross-lingual transfer. For completeness, we provide a brief overview of the framework in what follows.

MAD-X comprises three adapter types: language, task, and invertible adapters; this enables learning language and task-specific transformations in a modular and parameter-efficient way. As in prior work \cite{Rebuffi2017adapters,Houlsby2019adapters}, adapters are trained while keeping the parameters of the pretrained multilingual model fixed. \textit{Language adapters} are trained via masked language modeling (MLM) on unlabelled target language data. \textit{Task adapters} are trained via task-specific objectives on labelled task data in a source language while also keeping the language adapters fixed. Task and language adapters are stacked: this enables the adaptation of the pretrained multilingual model to languages not covered in its pretraining data. MAD-X keeps the same task adapter while substituting the source language adapter with the target language adapter at inference.

In brief, the adapters $\mathsf{A}_l$ at layer $l$ consist of a down-projection $\textbf{D} \in \mathbb{R}^{h \times d}$ where $h$ is the hidden size of the Transformer model and $d$ is the dimension of the adapter, followed by a $\mathsf{GeLU}$ activation \cite{Hendrycks:2020gelu} and an up-projection $\textbf{U}\in \mathbb{R}^{d \times h}$ at every layer $l$:
%
{
\begin{equation}
\mathsf{A}_l(\textbf{h}_l, \textbf{r}_l) = \textbf{U}_l(\mathsf{GeLU}(\textbf{D}_l(\textbf{h}_l))) + \textbf{r}_l.
\end{equation}
}%
\noindent $\textbf{h}_l$ and $\textbf{r}_l$ are the Transformer hidden state and the residual at layer $l$, respectively. The residual connection $\textbf{r}_l$ is the output of the Transformer's feed-forward layer whereas $\textbf{h}_l$ is the output of the subsequent layer normalization. For further technical details, we refer the reader to \newcite{pfeiffer20madx}.


Current model adaptation approaches \cite{ChauLS20Parsing,Wang20ExtendmBERT} generally fine-tune \textit{all} model parameters on target language data. Instead, we follow the more computationally efficient adapter-based paradigm where we keep model parameters fixed, and only train language adapters and target language embeddings. Crucially, while the current adapter-based methods offer extra capacity, they do not offer mechanisms to deal with extended vocabularies of many resource-poor target languages, and do not adapt their representation space towards the target language adequately. This problem is 
exacerbated when dealing with unseen languages and scripts.\footnote{An alternative approach based on transliteration \cite{Muller20BeingUnseen} side-steps script adaptation but relies on language-specific heuristics, which are not available for most languages.}

\begin{table}[!t]
\centering
\footnotesize
\def\arraystretch{0.99}
\resizebox{\columnwidth}{!}{
\begin{tabular}{llllcc}
\toprule
\multirow{2}{*}{Language} & \multirow{2}{*}{iso} & \multirow{2}{*}{Family} & \multirow{2}{*}{Script} & \% & \% Lex \\
& & & & \unk{}s & Overl.\\
\midrule
English* & en & Indo-Europ. & Latin & 0\% & 66\% \\
Chinese* & zh & Sino-Tibetan & Chinese & 0\% & 79\% \\
Japanese* & ja & Japonic & Japanese & 0\% & 99\% \\
Arabic* & ar & Afro-Asiatic & Arabic & 1\% & 6\% \\
Georgian & ka & Kartvelian & Georgian & 2\% & 27\% \\
Urdu & ur & Indo-Europ. & Arabic & 5\% & 34\% \\
Hindi & hi & Indo-Europ. & Devanagari & 2\% & 33\% \\ \cdashline{1-6} \vspace{-3mm} \\
Min Dong & cdo & Sino-Tibetan & Chinese & $6\%$ & $53\%$ \\
Māori & mi & Austronesian & Latin & $1\%$ & $45\%$ \\
Ilokano & ilo & Austronesian & Latin & $2\%$ & $48\%$ \\
Guarani & gn & Tupian & Latin & $3\%$ & $42\%$ \\
Mingrelian & xmf & Kartvelian & Georgian & 7\% & 22\% \\
Sindhi & sd & Indo-Europ. & Arabic & 30\% & 25\% \\
Erzya & myv & Uralic & Cyrilic & 1\% & 33\% \\
Bhojpuri & bh & Indo-Europ. & Devanagari & 1\% & 28\% \\
Wolof & wo & Niger-Congo & Latin & 1\% & 31\% \\ \cdashline{1-6} \vspace{-3mm} \\
Amharic & am & Afro-Asiatic & Ge'ez & 86\% & 13\% \\
Tibetan & bo & Sino-Tibetan & Tibetan & $66\%$ & $20\%$ \\
Khmer & km & Austroasiatic & Khmer & $79\%$ & $19\%$ \\
Divehi & dv & Indo-Europ. & Thaana & $85\%$ & $14\%$ \\
Sinhala & si & Indo-Europ. & Sinhala & $75\%$ & $23\%$ \\
\bottomrule
\end{tabular}
}
\caption{Languages in our NER evaluation, together with their language family and common script. For each monolingual vocabulary (\S\ref{sec:experimental_setup}), we compute the proportion of tokens that cannot be composed by the subword tokens from the original mBERT vocabulary (\unk{}s) as well as of \emph{lexically overlapping} tokens that are present in both vocabularies. *: For the high-resource source languages, proportions are calculated with regard to the tokenizers used by \citet{Rust2020}.
}  
\label{tab:ner-languages}
\end{table}

\section{Cross-lingual Transfer of Lexical Information} \label{sec:embedding-methods}

\begin{table*}[t!]
\centering
\def\arraystretch{0.99}
\resizebox{\textwidth}{!}{%
\begin{tabular}{lccccccc}
\toprule
\multirow{2}{*}{Method} & Special & Lexical & Latent semantic & Language & New & \# of new  & \multirow{2}{*}{Reference} \\
 & tokens & overlap & concepts & clusters & params & params   \\ \midrule
\elrand & \checkmark &  &  & & $\mathbf{X'}$ & 7.68M   & \citet{Artetxe2020cross-lingual} \\ 
\ellex & \checkmark & \checkmark &  & & $\mathbf{X'}$ & 7.68M   & \citet{ChauLS20Parsing,Wang20ExtendmBERT} \\ 
\textsc{MF}$^{C}_{*}$-\textsc{rand} & \checkmark & & \checkmark & \checkmark & $\mathbf{F'}$, $\mathbf{I'}$ & 1M + $C\cdot $10k    & Ours \\
\textsc{MF}$^{C}_{*}$-\textsc{lex} & \checkmark & \checkmark & \checkmark & \checkmark & $\mathbf{F'}$, $\mathbf{I'}$ & 1M + $C\cdot $10k    & Ours \\
\bottomrule
\end{tabular}%
}
\caption{Overview of our methods and related approaches together with the pretrained knowledge they utilize. We calculate the number of new parameters per language with $V' = 10k$, $D=768$, and $D'= 100$. We do not include up-projection matrices $\mathbf{G}$ as these are learned only once and make up a comparatively small number of parameters.}
\label{tab:embedding-adaptation-approaches}
\end{table*}

The embedding matrix of large multilingual models makes up around 50\% of their entire parameter budget \cite{Chung2020rethinking}. However, it is not clear how to leverage this large amount of information most effectively for languages that are not adequately represented in the shared multilingual vocabulary due to lack of pretraining data.

\begin{figure}[!t]
        \centering
        \includegraphics[width=0.95\linewidth]{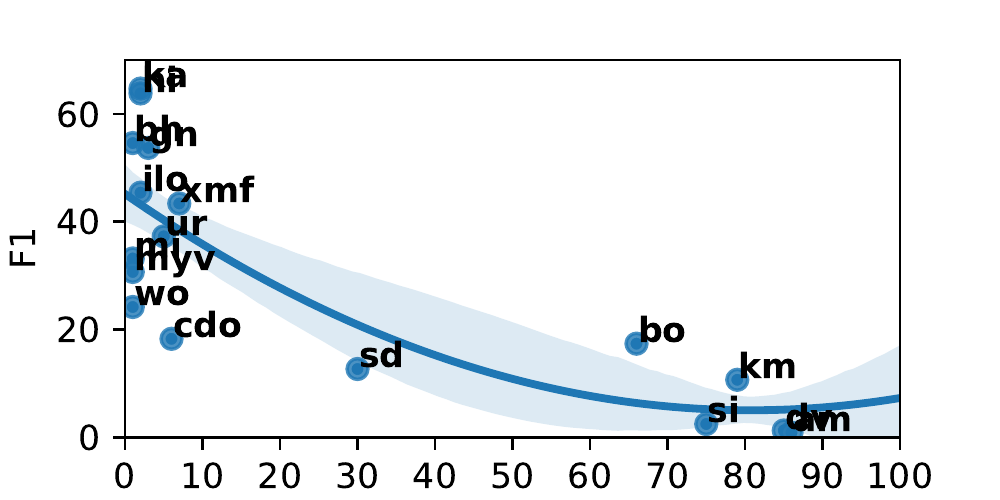}
    \vspace{-1mm}
    \caption{mBERT's zero-shot cross-lingual transfer performance with respect to the proportion of  \unk{}s in mBERT's original vocabulary relative to the target language (see Table~\ref{tab:ner-languages}).}
    \vspace{-1mm}
\label{fig:scatter_mBERT_UNKs}
\vspace{-1.5mm}
\end{figure}

A key challenge in using the lexical information encoded in the embedding matrix is to overcome a mismatch in vocabulary between the pretrained model and the target language. To outline this issue, in Table~\ref{tab:ner-languages} we show for the languages in our NER evaluation the proportion of tokens in each language that are effectively \textit{unknown} (\unk) to mBERT: they occur in the vocabulary of a separately trained monolingual tokenizer (\textsection \ref{sec:experimental_setup}), but cannot even be composed by subword tokens from the original mBERT's vocabulary. Table~\ref{tab:ner-languages} also provides the proportion of \textit{lexically overlapping} tokens, i.e., tokens that are present both in mBERT's and monolingual in-language vocabularies. The zero-shot performance of mBERT generally deteriorates with less lexical overlap and more \unk{}s in a target language: see 
Figure~\ref{fig:scatter_mBERT_UNKs}. Pearson's $\rho$ correlation scores between the lexical overlap and proportion of \unk{}s (see Table~\ref{tab:ner-languages}) and NER performance are $0.443$ and $-0.798$, respectively.

Recent approaches such as invertible adapters \cite{pfeiffer20madx} that adapt embeddings in the pretrained multilingual vocabulary may be able to deal with lesser degrees of lexical overlap. Still, they cannot deal with \unk tokens. In the following, we systematize existing approaches and present novel ways of adapting a pretrained model to the vocabulary of a target language that can handle this challenging setting, present most acutely when adapting to languages with unseen scripts. We summarize the approaches in Table~\ref{tab:embedding-adaptation-approaches} based on what types of pretrained information they utilize. 
All approaches rely on a new vocabulary $V'$, learned on the target language data.

\subsection{Target-Language Embedding Learning}
\label{sec:tlel}
A straightforward way to adapt a pretrained model to a new language is to learn new embeddings for the language. Given the new vocabulary $V'$, we initialize new embeddings $\mathbf{X}' \in \mathbb{R}^{|V'| \times D}$ for all $V'$ vocabulary items where $D$ is the dimensionality of the existing embeddings $\mathbf{X} \in \mathbb{R}^{|V| \times D}$, and only initialize special tokens (e.g. \texttt{[CLS]}, \texttt{[SEP]}) with their pretrained representations. We train the new embeddings of the $\mathbf{X}'$ with the pretraining task. This approach, termed \elrand, was proposed by \citet{Artetxe2020cross-lingual}: they show that it allows learning aligned representations for a new language but only evaluate on high-resource languages. The shared special tokens allow the model to access a minimum amount of lexical information, which can be useful for transfer \cite{Dufter2020elements}. Beyond this, this approach leverages knowledge from the existing embedding matrix only implicitly to the extent that the higher-level hidden representations are aligned to the lexical representations. 

\subsection{Initialization with Lexical Overlap}
\label{sec:initlex}
 
To leverage more lexical information, we can apply shared initialization not only to the special tokens but to all lexically overlapping tokens. Let us denote this vocabulary subset with $V'_{lex}$, and $V'_{rand} = V' \setminus V'_{lex}$. In particular, we initialize the embeddings of all lexically overlapping tokens $\mathbf{X}'_{lex}$ from $V'_{lex}$ with their pretrained representations from the original matrix $\mathbf{X}$, while the tokens from $V'_{rand}$ receive randomly initialized embeddings $\mathbf{X}'_{rand}$. We then fine-tune all target language embeddings $\mathbf{X}' = \mathbf{X}'_{lex} \cup \mathbf{X}'_{rand}$ on the target language data.
\citet{Wang20ExtendmBERT} cast this as extending $V$ with new tokens. In contrast, we seek to disentangle the impact of vocabulary size and pretrained information. 
As one variant of this  approach, \citet{ChauLS20Parsing} only add the 99 most common tokens of a new language to $V$.
 
Initialization with lexical overlap, termed \ellex, allows us to selectively leverage the information from the pretrained model on a per-token level based on surface-level similarity. Intuitively, this should be most useful for languages that are lexically similar to those seen during pretraining and have a substantial proportion of lexically overlapping tokens. However, such lexical overlap is a lot rarer for languages that are written in different scripts. For such languages, relying on surface-level string similarity alone may not be enough.

\subsection{Embedding Matrix Factorization} 
We therefore propose to identify latent semantic concepts in the pretrained model's embedding matrix that are general across languages and useful for transfer. 
Further, to allow modeling flexibility we propose to learn a grouping of similar tokens. We achieve this by factorizing the pretrained embedding matrix $\textbf{X} \in \mathbb{R}^{|V| \times D}$ into lower-dimensional word embeddings $\textbf{F} \in \mathbb{R}^{|V| \times D'}$ and $C$ shared up-projections $\textbf{G}^1, \ldots, \textbf{G}^C \in \mathbb{R}^{D' \times D}$ that encode general cross-lingual information: 
{
\begin{equation}
\textbf{X} \approx  \sum_{c \in C}  \text{diag}(\textbf{i}_c) \textbf{F} \textbf{G}^c
    \label{eq:FIG}
\end{equation}
}%
\noindent $D'$ is the dimensionality of the lower-dimensional embeddings. $\textbf{I} = [\textbf{i}_1, \ldots, \textbf{i}_C] \in \mathbb{R}^{|V| \times C}$ is an indicator matrix where $i_{v, c} = 1$ iff token $v$ is associated with the $c$-th up-projection, else $0$.\footnote{We use bold upper case letters ($\mathbf{X}$) for matrices, subscripts with bold letters ($\mathbf{x}_i$) for rows or columns and subscripts with standard weight letters for specific elements ($x_i$, $x_{i, j}$).} $\text{diag}(\cdot)$ creates a diagonal matrix from a vector, i.e. $\text{diag}(\textbf{i}_c) \in \mathbb{R}^{|V| \times |V|}$.

\vspace{1.3mm}
\noindent \textbf{\textsc{MF$^{1}$-$*$}.} In the basic case where $C =  1$, Eq.~(\ref{eq:FIG}) simplifies to $\textbf{X} \approx  \textbf{FG}$.
As $\textbf{X}$ is unconstrained, we follow \citet{ding2008convex} and interpret this as a semi-non-negative matrix factorization (Semi-NMF) problem. In Semi-NMF, $\textbf{G}$ is restricted to be non-negative while no restrictions are placed on the signs of $\textbf{F}$. The Semi-NMF is computed via an iterative updating algorithm that alternatively updates $\textbf{F}$ and $\textbf{G}$ where the Frobenius norm is minimized.\footnote{For more details see \citet{ding2008convex}.}  

$\textbf{G}$ is shared across all tokens and thus encodes general properties of the original embedding matrix $\textbf{X}$ whereas $\textbf{F}$ stores token-specific information.  $\textbf{G}$ only needs to be pretrained once and can be used and fine-tuned for every new language. To this end, we simply learn new low-dimensional embeddings $\textbf{F}' \in \mathbb{R}^{|V'| \times D'}$ with the pretraining task, which are up-projected with $\textbf{G}$ and fed to the model. 

\vspace{1.3mm}
\noindent \textbf{\textsc{MF}$^C_{\textsc{KMeans}}$-$*$.}
When $C>1$, each token is associated with one of $C$ up-projection matrices. Grouping tokens and using a separate up-projection matrix per group may help balance sharing information across typologically similar languages with learning a robust representation for each token \cite{ChungGTR20}. We propose two approaches to automatically learn such a clustering. 
 
In our first, pipeline-based approach, we first cluster $\textbf{X}$ into $C$ clusters  using KMeans. For each cluster, we then factorize the subset of embeddings $\textbf{X}^c$ associated with the $c\text{-th}$ cluster separately using Semi-NMF equivalently as for \textsc{MF$^{1}$-$*$}.

For a new language, we learn new low-dim embeddings $\textbf{F}' \in \mathbb{R}^{|V'| \times D'}$ and a randomly initialized matrix $\textbf{Z} \in  \mathbb{R}^{|V'| \times C}$, which allows us to compute the cluster assignment matrix $\textbf{I}' \in  \mathbb{R}^{|V'| \times C}$. Specifically, for token $v$, we obtain its cluster assignment as $\argmax$ of $\textbf{z}_{v,\cdot}'$. 
As $\argmax$ is not differentiable, we employ the Straight-Through Gumbel-Softmax estimator \cite{Jang2017} defined as:
{
\begin{equation}
i'_{v,c} = \frac{\exp(\text{log}(z_{v,c}) + g_c)/\tau)}{\sum^C_{j=1} \exp(\text{log}(z_{v,j}) + g_j)/\tau)},
\label{eq:gumbel}
\end{equation}
}%
\noindent where  $\tau$ is a temperature parameter, and  $\mathbf{g} \in \mathbb{R}^{|V|}$ corresponds to samples from the Gumbel distribution $g_j \sim - \log(- \log(u_j))$
with $u_j \sim \mathcal{U}(0, 1)$ being the uniform distribution. $\textbf{z}_{v,\cdot}$ can be seen as ``logits'' used for assigning the $v$-th token a cluster. As $\tau \rightarrow 0$, the softmax becomes an $\argmax$ and the Gumbel-Softmax distribution approximates more closely the categorical distribution. $\textbf{I}' \in \mathbb{R}^{|V'| \times C}$ represents the one-hot encoded, indicator function over possible clusters, with learnable parameters $\textbf{Z}$. As before, $i'_{v, c} = 1$ iff new token $v$ is associated with up-projection $c$, else $0$.

\vspace{1.3mm}
\noindent \textbf{\textsc{MF}$^C_{\textsc{Neural}}$-$*$.} We can also learn the cluster assignment and up-projections jointly. Specifically, we parameterize $\mathbf{G}$ in Eq.~\eqref{eq:FIG} using a neural net where we learn the indicator matrix $\textbf{I}$ equivalently to Eq.~\eqref{eq:gumbel}. 
The objective minimizes the L$_2$-norm between the original and predicted embeddings:
{
\begin{equation}
        \mathcal{L} = ||\textbf{X} - \sum_{c \in C}  \text{diag}(\mathbf{i}_{c}) \textbf{F} \textbf{G}^c||_2
    \label{eq:l2loss}
\end{equation}
}%
\noindent For a new language, we proceed analogously. 

\vspace{1.3mm}
\noindent \textbf{\textsc{MF$^{*}_{*}$-\textsc{rand}} and \textsc{MF$^{*}_{*}$-\textsc{lex}}.}
Finally, we can combine different initialization strategies (see \S\ref{sec:tlel} and \S\ref{sec:initlex}) with the embedding matrix factorization technique. We label the variant which relies on random initialization, see \S\ref{sec:tlel}, as \mfrand. The variant, which relies on lexically overlapping tokens from \S\ref{sec:initlex} can leverage both surface-level similarity as well as latent knowledge in the embedding matrix; we simply initialize the embeddings of overlapping lexical tokens (from $V'_{lex}$) in $\textbf{F}'$ with their low-dim representations from $\textbf{F}$. The remaining tokens (from $V'_{rand}$) are randomly initialized in $\textbf{F}'$.

Factorizing the embedding matrix has the additional benefit of reducing the number of trainable parameters and correspondingly the amount of storage space required for each additional language. This is especially true when $D \gg D'$.\footnote{For instance, when $|V| = |V|' = 10k$, $D = 768$,  and $D' = 100$ as in our experiments we reduce the amount of storage space required by $82\%$ per language.}

\section{Experiments}
\noindent \textbf{Data.}\hspace{0.3mm} 
For pretraining, we leverage the Wikipedia dumps of the target languages. 
We conduct experiments on 
named entity recognition (NER) and dependency parsing (DP). 
For NER, we use the WikiAnn \cite{Pan2017wikiann} dataset, partitioned into train, dev, and test portions by \citet{Rahimi2019massively}. For DP we use Universal Dependencies \citep[UD;][]{nivre:2016, nivre:2020, zeman:2020}. 

\vspace{1.3mm}
\noindent \textbf{Languages.}\hspace{0.3mm} 
WikiAnn offers a wide language coverage (176 languages) and, consequently, a number of language-related comparative analyses. In order to systematically compare against state-of-the-art cross-lingual methods under different evaluation conditions, we identify \textbf{1)} low-resource languages where the script has been covered by mBERT but the model has not been specifically trained on the language \textbf{2)} as well as low-resource languages with scripts not covered at all by pretraining. In each case, we select four languages, taking into account variance in data availability and typological diversity. We select four high-resource source languages (English, Chinese, Japanese, Arabic) in order to go beyond English-centric transfer. We evaluate the cross-lingual transfer performance of these 4 languages to the 17 diverse languages. For DP, we chose the subset that occurs in UD. We highlight the properties of all 21 languages in Table~\ref{tab:ner-languages}.

\vspace{-0.3em}
\subsection{Baselines}
\vspace{-0.1em} 

\noindent \textbf{mBERT (Standard Transfer Setup).}\hspace{0.3mm}
We primarily focus on mBERT as it has been shown to work well for low-resource languages \cite{pfeiffer20madx}. mBERT is trained on the 104 languages with the largest Wikipedias.\footnote{See Appendix Table~\ref{table:list_of_languages}  for the list of all 104 covered languages with corresponding scripts.} 
In the standard cross-lingual transfer setting (see \S\ref{s:rw}), the full model is fine-tuned on the target task in the (high-resource) source language, and is evaluated on the test set of the target (low-resource) language. 

\vspace{1.3mm}
\noindent {\bf \madx.}\hspace{0.3mm} 
We follow \citet{pfeiffer20madx} and stack task adapters on top of pretrained language adapters (see \S\ref{s:rw}). When training the model on source language task data, only the task adapter is trained while the original model weights and the source language adapter are frozen. At inference, the source language adapter is replaced with the target language adapter.

\vspace{1.3mm}
\noindent {\bf \madxtwo.}\hspace{0.3mm}
The adapter in the last transformer layer is not encapsulated between frozen transformer layers, and can thus be considered an extension of the prediction head. This places no constraints on the representation of the final adapter, possibly decreasing transfer performance when replacing the language adapters for zero-shot transfer. In this work, we thus propose to drop the adapter in the last transformer layer, and also evaluate this novel variant of the MAD-X framework.

\subsection{Methods}
\label{sec:methods} 

We experiment with the methods from Table~\ref{tab:embedding-adaptation-approaches} and discussed in \textsection \ref{sec:embedding-methods}, summarized here for clarity.

\vspace{1mm}
\noindent \textbf{\textsc{EL-*}}\hspace{0.3mm} 
We randomly initialize embeddings for all tokens---except special tokens---in the new vocabulary (\elrand) or initialize embeddings of lexically overlapping tokens with their pretrained representations (\ellex).  

\vspace{1mm}
\noindent \textbf{\textsc{MF$^1$-*}}\hspace{0.3mm} We randomly initialize lower-dimensional embeddings (\mfrand) or initialize lexically overlapping tokens with their corresponding lower-dimensional pretrained representation (\mflex) while using a single pretrained projection matrix. 

\vspace{1.3mm}
\noindent \textbf{\textsc{MF$^{10}_{*}$-*}}\hspace{0.3mm} We learn assignments to 10 clusters via k-means and up-projection matrices via Semi-NMF (\textsc{MF$^{10}_{\textsc{KMeans}}\text{-}*$}). Alternatively, we learn cluster assignments with Gumbel-Softmax and up-projection matrices jointly
(\textsc{MF$^{10}_{\textsc{Neural}}\text{-}*$}). For new tokens we use random (\textsc{MF$^{10}_{*}$-rand}) or lexical overlap initialisation (\textsc{MF$^{10}_{*}\text{-lex}$}).

\begin{table*}[!t]

\begin{subtable}[t]{\linewidth}
\centering
\def\arraystretch{0.99}
\footnotesize
\resizebox{\textwidth}{!}{
\begin{tabular}{lrrrr|rrrrrrrrrr|rrrrrr|r}
\toprule
    & \multicolumn{4}{c}{Seen Languages} & \multicolumn{10}{c}{Unseen Languages but Covered Scripts} & \multicolumn{6}{c}{New Scripts} & \textbf{\textit{Mac.}} \\ 

	 & 	ka & ur & hi & \textbf{Avg} & cdo & mi & ilo & gn & xmf & sd & myv & bh & wo & \textbf{Avg} & am & bo & km & dv & si & \textbf{Avg} &  \textbf{\textit{Avg}} \\
\midrule
mBERT & 64.7 & 37.3 & \textbf{63.8} & 55.3 & 18.3 & 33.2 & 45.4 & 53.7 & 43.4 & 12.7 & 30.7 & 54.6 & 24.2 &  35.1 & 0.9 & 17.4 & 10.7 & 1.3 & 2.5 & 6.6 & \textit{30.3} \\
\textsc{MAD-X}	 & 	63.9 & 51.4 & 58.7 & 58.0 & 29.1 & 45.6 & 45.5 & 52.7 & 51.9 & 34.0 & 57.8 & 57.0 & 49.5 & 47.0 & \textbf{10.8} & \textbf{24.8} & 17.6 & 16.8 & 16.8 & 17.4  & \textit{40.2} \\
\textsc{MAD-X 2.0} & \textbf{65.6} & \textbf{55.3} & 61.0 & \textbf{60.6} & \textbf{30.7} & \textbf{50.6} & \underline{\textbf{64.0}} & \underline{\textbf{56.3}} & \textbf{52.6} & \textbf{37.1} & \underline{\textbf{63.0}} & \textbf{59.8} & \textbf{55.5} & \textbf{52.2} & 10.7 & 24.7 & \textbf{18.1} & \textbf{22.2} & \textbf{18.7} & \textbf{18.9} & \textbf{\textit{43.9}}  \\
\midrule
\elrand	 & 	\textbf{65.8} & 47.8 & \textbf{63.8} & \textbf{59.1} & 38.0 & 7.3 & \textbf{57.9} & 48.5 & 59.4 & 44.2 & 35.2 & 55.5 & 5.1 & 39.0 & \textbf{42.9} & \underline{\textbf{53.9}} & 59.5 & 32.7 & 51.5 & 48.1 & \textit{45.2} \\
\mfrand & 63.5 & 48.5 & 57.8 & 56.6  & 39.8 & \textbf{19.3} & 43.6 & 47.3 & 57.9 & 45.1 & \textbf{60.4} & 44.4 & \textbf{45.0} & \textbf{44.8} & 42.6 & 49.0 & 61.7 & \textbf{43.4} & \textbf{52.1} & \textbf{49.8} & \textbf{\textit{48.3}} \\
\mfrandkmeans & 64.3 & \textbf{52.6} & 60.4 & \textbf{59.1} & 20.7 & 11.7 & 51.2 &	50.8& 58.2 & \textbf{45.2} & 49.0 & \textbf{60.0} & 42.5 & 43.2 & 37.3 & 37.5 & \textbf{64.2} & 18.5 & 47.2 & 40.9 & \textit{45.4}\\
\mfrandnn & 63.2 & 49.4 & 60.6 & 57.7 & \textbf{40.7} & 5.5 & 48.5 & \textbf{52.0} & \underline{\textbf{60.4}} & 27.2 & 46.7 & 54.5 & 41.2 & 41.9 & 42.7 & 38.4 & 63.7 & 38.7 & 48.0 & 46.3 & \textit{46.0} \\
\cdashline{1-22} \vspace{-2mm} \\
\ellex & \underline{\textbf{69.3}} & 57.9 & \underline{\textbf{66.4}} & \underline{\textbf{64.5}} & 46.8 & 32.2 & \textbf{58.8} & 54.1 & 57.0 & 44.6 & \textbf{56.3} & 59.5 & 50.1 & 51.0 & \underline{\textbf{46.5}} & 51.5 & 61.0 & \underline{\textbf{47.2}} & \underline{\textbf{56.3}} & \underline{\textbf{52.5}} & \textit{53.6} \\
\mflex & 65.5 & 54.0 & 61.5 & 60.4  & 46.6 & 39.8 & 55.2 & \textbf{54.8} & \textbf{57.2} & 45.0 & 55.0 & 59.1 & 48.1 & 51.2 & 38.6 & 42.5 & 63.9 & 43.8 & 51.5 & 48.0 & \textit{51.2}\\
\mflexkmeans & 66.3 & \underline{\textbf{59.5}} & 61.1 & 62.3 & \underline{\textbf{48.9}} & 47.5 & 46.4 & 53.6 & 56.7 & 46.2 & 58.4 & \underline{\textbf{61.0}} & \underline{\textbf{57.9}} & \underline{\textbf{53.0}} & 45.5 & \textbf{51.6} & 64.2 & 44.9 & 46.2 & 50.5 &  \underline{\textbf{\textit{53.9}}} \\
\mflexnn & 67.0 & 55.8 & 62.6 & 61.8 & 47.8 & \underline{\textbf{50.8}} & 53.1 & 53.9 & 55.6 & \underline{\textbf{46.4}} & 50.4 & 60.6 & 51.8 & 52.3  & 46.7  & 43.0  & \underline{\textbf{65.2}}  & 46.3  & 51.5  & 50.5 & \textit{53.4}\\
\bottomrule
\end{tabular}
}

\caption{ Named Entity Recognition: Mean F$_1$ test results for UD averaged over 5 runs and averaged over the 4 source languages. 
}
\label{tab:mainresults_ner}
\end{subtable}

\vspace{1em}

\begin{subtable}[t]{\linewidth}
%
\centering
\def\arraystretch{0.99}
\footnotesize
\resizebox{\textwidth}{!}{
\begin{tabular}{lrrr|rrrr|r|r}
\toprule
    & \multicolumn{3}{c}{Seen Languages} & \multicolumn{4}{c}{Unseen Languages but Covered Scripts} & \multicolumn{1}{c}{New Script} &  \\ 
	 & 	 \multicolumn{1}{c}{hi} & \multicolumn{1}{c}{ur} & \multicolumn{1}{c}{\textbf{Avg}} & \multicolumn{1}{c}{bh} & \multicolumn{1}{c}{myv} &  \multicolumn{1}{c}{wo} & \multicolumn{1}{c}{\textbf{Avg}} & \multicolumn{1}{c}{am} &\multicolumn{1}{c}{\textbf{\textit{Macro Avg}}}  \\
		 & 	 \multicolumn{1}{c}{UAS / LAS} & \multicolumn{1}{c}{UAS / LAS} & \multicolumn{1}{c}{UAS / LAS} & \multicolumn{1}{c}{UAS / LAS} & \multicolumn{1}{c}{UAS / LAS} &  \multicolumn{1}{c}{UAS / LAS} & \multicolumn{1}{c}{UAS / LAS} & \multicolumn{1}{c}{UAS / LAS}  & \multicolumn{1}{c}{UAS / LAS}  \\
\midrule
mBERT & \underline{\textbf{45.8}} / \underline{\textbf{29.5}} & 34.9 / \textbf{20.5} & \underline{\textbf{40.3}} / \underline{\textbf{25.0}} & \underline{\textbf{33.9}} / \underline{\textbf{18.6}} & 29.7 / 13.0 & 27.5 /  \, 7.5 & 30.4 / 13.0 & 10.1 /  \, 3.8 & \textit{30.3 / 15.5}\\

\textsc{MAD-X 2.0} & 42.0 / 26.1 & \textbf{35.0} / 20.1 & 38.5 / 23.1 & 31.0 / 16.2 & \textbf{47.6} / \underline{\textbf{29.8}} & \textbf{35.4} / \textbf{18.4} & \textbf{38.0} / \textbf{21.4} & \textbf{14.0} /  \, \textbf{7.7} & \textit{\textbf{34.2} / \textbf{19.7}} \\
\midrule
\elrand	 & 	41.9 / 25.9 & 32.9 / 18.4 & 37.4 / 22.1 & 29.9 / 13.8 & 45.4 / 23.9 & 24.3 /  \, 7.2 & 33.2 / 15.0 & 30.1 / 11.3 & \textit{34.1 / 16.7}\\
\mfrand & 41.7 / 26.2 & 33.7 / 19.4 & 37.7 / 22.8 & 28.7 / 13.7 & 47.6 / 27.2 & \textbf{32.7} / \textbf{15.5} & 36.3 / 18.8 & \textbf{34.3} / \textbf{13.3} & \textbf{36.4} / \textbf{19.2}\\
\mfrandkmeans  & 40.7 / 25.1 & 30.4 / 17.7 & 35.6 / 21.4 & \textbf{31.9} / \textbf{16.3} & 47.8 / 27.7 & 29.6 / 13.7 & \textbf{36.4} / \textbf{19.2} & 22.6 / 10.1 & \textit{33.8 / 18.4}\\
\mfrandnn & \textbf{43.1} / \textbf{27.0} & \textbf{34.7} / \textbf{20.0} & \textbf{38.9} / \textbf{23.5} & 30.8 / 14.7 & \textbf{48.6} / \textbf{28.7} & 21.7 / \, 9.4 & 33.7 / 17.6 & 32.4 / 12.4 & \textit{35.2 / 18.7} \\
\cdashline{1-10} \vspace{-2.5mm} \\
\ellex & 41.9 / 26.1 & 34.0 / 19.6 & 37.9 / 22.9 & 30.6 / 16.2 & 48.8 / 27.7 & 37.9 / 19.2 & 39.1 / 21.0 & 34.0 /  \, 9.9 & \textit{36.6 / 19.8}\\
\mflex & 42.1 / 26.7 & 34.0 / 20.1 & 38.0 / 23.4 & \textbf{31.2} / \textbf{16.4} & \underline{\textbf{49.0}} / \textbf{28.9} & 37.6 / 19.5 & \underline{\textbf{39.3}} / \underline{\textbf{21.6}} & \underline{\textbf{34.8}} / \underline{\textbf{12.9}} & \textit{36.8 / \underline{\textbf{20.8}}} \\
\mflexkmeans  & 42.2 / 27.0 & 34.7 / 20.3 & 38.5 / 23.6 & 31.1 / \textbf{16.4} & 47.7 / 28.6 & 33.8 / 17.3 & 37.5 / 20.8 & 32.2 / 13.2 & \textit{37.0 / 20.5} \\
\mflexnn & \textbf{43.0} / \textbf{27.4} & \underline{\textbf{36.1}} / \underline{\textbf{21.2}} & \textbf{39.6} / \textbf{24.3} & 30.9 / 16.2 & 46.8 / 27.6 & \underline{\textbf{38.7}} / \underline{\textbf{20.4}} & 38.8 / 21.4 & 31.4 / 12.1 & \underline{\textit{\textbf{37.8 / 20.8}}} \\
\bottomrule
\end{tabular}
}
\caption{ Dependency Parsing: Mean UAS and LAS test results for UD averaged over 5 runs and averaged over the 4 source languages.  
}
\label{tab:mainresults_UD}
\end{subtable}
\caption{ Mean test results for (a) NER and (b) DP.  
The top group (first two/three rows) includes models, which leverage the original tokenizer which is not specialized for the target language. The second group (last eight rows) include models with new tokenizers. We separate models with randomly initialized embeddings ($*$-\textsc{RandInit}) from models with lexical init ($*$-\textsc{LexInit}) by the dashed line. Bold numbers indicate best-scoring models of the respective group, underlined numbers the best performance overall. Source languages are \textit{en}, \textit{ar}, \textit{zh}, and \textit{ja}.}
\label{tab:mainresults}
\end{table*}
 
\subsection{Experimental Setup}
\label{sec:experimental_setup}
Previous work generally fine-tunes the entire model on the target task \cite{ChauLS20Parsing,Wang20ExtendmBERT}. To extend the model to a new vocabulary, \citet{Artetxe2020cross-lingual} alternatingly freeze and fine-tune embeddings and transformer weights for pretraining, and target task fine-tuning, respectively. We find that this approach largely underperforms adapter-based transfer as proposed by \citet{pfeiffer20madx}, and we thus primarily focus on adapter-based training in this work.\footnote{We present results with such full model transfer in \S\ref{sec:ap_full_model_transfer}.}

\vspace{1.3mm}
\noindent \textbf{Adapter-Based Transfer.}\hspace{0.3mm}
We largely follow the experimental setup of \citet{pfeiffer20madx}, unless noted otherwise. We obtain language adapters for the high-resource languages from \href{https://AdapterHub.ml}{AdapterHub.ml} \cite{pfeiffer2020AdapterHub} and train language adapters and embeddings for the low-resource languages jointly while keeping the rest of the model fixed. For zero-shot transfer, we replace the source language adapter with the target adapter, and also replace the entire embedding layer with the new embedding layer specialized to the target language. \madxtwo consistently outperforms \madx (see \S\ref{sec:results}); we thus use this setup for all our methods.  

\vspace{1.3mm}
\noindent \textbf{Tokenizer.}\hspace{0.3mm} 
We learn a new WordPiece tokenizer for each target language with a vocabulary size of 10k using the HuggingFace tokenizer library.\footnote{\href{https://github.com/huggingface/tokenizers}{https://github.com/huggingface/tokenizers}} 

\vspace{1.3mm}
\noindent \textbf{Semi-NMF.}\hspace{0.3mm}
We factorize the pretrained embedding matrix of mBERT using Semi-NMF \cite{ding2008convex} leveraging the default implementation provided by \citet{bauckhagefactorizing}.\footnote{\href{ https://github.com/cthurau/pymf }{ https://github.com/cthurau/pymf }} We train for 3,000 update steps and leverage the corresponding matrices $\textbf{F}$ and $\textbf{G}$ as initialization for the new vocabulary. We choose the reduced embedding dimensionality $D'=100$.  $\textbf{F}$ is only used when initializing the (lower-dimensional) embedding matrix with lexically overlapping representations. 

\vspace{1mm}
\noindent \textbf{Masked Language Modeling.}\hspace{0.3mm}
For MLM pretraining we leverage the entire Wikipedia corpus of the respective language. We train for 200 epochs or $\sim$100k update steps, depending on the corpus size. The batch size is 64; the learning rate is $1e-4$.

\vspace{1.3mm}
\noindent \textbf{Task Fine-tuning.}\hspace{0.3mm}
Our preliminary experiments suggested that fine-tuning the model for a smaller number of epochs leads to better transfer performance in low-resource languages in general. We thus fine-tune all the models for 10 epochs, evaluating on the source language dev set after every epoch. We then take the best-performing model according to the dev $F_1$ score, and use it in zero-shot transfer. We train all the models with a batch size of 16 on high resource languages. For NER we use learning rates $2e-5$ and $1e-4$ for full fine-tuning and adapter-based training, respectively. For DP, we use a transformer-based variant \cite{glavas:2020} of the standard deep biaffine attention dependency parser \cite{dozat:2017} and train with learning rates $2e-5$ and $5e-4$ for full fine-tuning and adapter-based training respectively.

\section{Results and Discussion} \label{sec:results}
 
\begin{table*}
\centering
\footnotesize
\resizebox{\textwidth}{!}{
\begin{tabular}{lrrr|rrrrrrrrr|rrrrr}
\toprule
            &   ka & ur & hi  & cdo  & mi   & ilo  & gn & xmf & sd & myv & bh & wo & am  & bo & km   & dv   & si   \\
\midrule
Numbers     & 11\%   & 8\%  & 6\%  & 26\% & 17\% & 8\%  & 8\%  & 15\% & 9\%  & 11\% & 11\% & 7\%  & 13\% & 9\%  & 3\%  & 12\% & 9\%  \\
Lat Char    & 4\%    & 3\%  & 3\%  & 2\%  & 2\%  & 2\%  & 2\%  & 5\%  & 4\%  & 3\%  & 4\%  & 3\%  & 8\%  & 5\%  & 5\%  & 7\%  & 4\%  \\
Lat (S)W    & 10\%   & 9\%  & 7\%  & 54\% & 61\% & 64\% & 65\% & 10\% & 10\% & 9\%  & 26\% & 71\% & 19\% & 30\% & 31\% & 17\% & 30\% \\
Oth Char    & 55\%   & 36\% & 48\% & 18\% & 19\% & 26\% & 25\% & 60\% & 39\% & 26\% & 24\% & 18\% & 60\% & 52\% & 61\% & 63\% & 57\% \\
Oth (S)W    & 20\%   & 43\% & 36\% & 0\%  & 1\%  & 0\%  & 0\%  & 10\% & 38\% & 51\% & 35\% & 1\%  & 0\%  & 3\%  & 0\%  & 1\%  & 0\% \\
\bottomrule
\end{tabular}
}
\caption{Grouping of tokens that lexically overlap between the original mBERT tokenizer and the tokenizer of the target language. \textit{Numbers} includes all tokens which include at least one number; \textit{Lat Char} indicates all Latin tokens of length 1; \textit{Lat (S)W} includes all (sub)words that include a Latin character but are of length $>$ 1. Consequently, \textit{Oth Char} and \textit{Oth (S)W} consists of characters and (sub)words respectively, which do not include Latin characters. }
\label{tab:lex-overlap-groupings}
\end{table*}

\begin{table*}[t!]
\centering
\resizebox{\textwidth}{!}{%
\begin{tabular}{llll|llll}
\toprule
cdo  & mi   & ilo  & gn   & bo & km   & dv   & si \\
\midrule
Northumberland	& Massachusetts	&	Melastomataceae	& establecimiento	&   University	&	languages	&	government	& International		\\
Massachusetts	& Encyclopedia	&	munisipalidad	& vicepresidente	&   Therefore	&	language	&	Chinese		& Bangladesh		\\
International	& Pennsylvania	&	Internasional	& Internacional		&   suffering	&	formula		&	govern		& wikipedia			\\
Pennsylvania	& Jacksonville	&	internasional	& internacional		&   existence	&	disease		&	system		& Australia			\\
Philadelphia	& Turkmenistan	&	International	& Independencia		&   practice	&	control		&	ation		& Zimbabwe			\\

\bottomrule
\end{tabular}
}
\caption{Longest lexically overlapping (sub)words.} 
\label{table:top_lex_overlap}
\end{table*}

The main results are summarised in Table~\ref{tab:mainresults_ner} for NER, and in Table~\ref{tab:mainresults_UD} for DP. 
First, our novel \madxtwo considerably outperforms the \madx version of \citet{pfeiffer20madx}. However, while both \madx versions improve over mBERT for unseen scripts, the performance remains quite low on average.  
The corresponding mBERT tokenizer is not able to adequately represent unseen scripts: many tokens are substituted by \unk{s}  (see Table~\ref{tab:ner-languages}), culminating in the observed low performance.

For our approaches that learn new embedding matrices, we observe that for languages seen during pretraining, but potentially underrepresented by the model (e.g. Georgian (ka), Urdu (ur), and Hindi (hi)), the proposed methods outperform \madxtwo for all tasks. This is in line with contemporary work \cite{Rust2020}, which emphasizes the importance of tokenizer quality for the downstream task. 
Consequently, for unseen languages with under-represented scripts, the performance gains are even larger, e.g., we see large improvements for Min Dong (cdo), Mingrelian (xmf), and Sindhi (sd). For unseen languages with the Latin script, our methods perform competitively (e.g. Maori (mi), Ilokano (ilo), Guarani (gn), and Wolof (wo)): this empirically confirms that the Latin script is adequately represented in the original vocabulary. The largest gains are achieved for languages with unseen scripts (e.g. Amharic (am), Tibetan (bo), Khmer (km), Divehi (dv), Sinhala (si)), as these languages are primarily represented as \unk{} tokens by the original mBERT tokenizer.

We observe improvements for most languages  
with lexical overlap initialization.
This adds further context to prior studies which found that a shared vocabulary is not necessary for learning multilingual representations \cite{Conneau:2020acl,Artetxe2020cross-lingual}: while it is possible to generalize to new languages without lexical overlap, leveraging the overlap still offers additional gains.

The methods based on matrix factorization (\textsc{MF$_*^*\text{-}*$})  improve performance over full-sized embedding methods (\textsc{EL$_*^*\text{-}*$}), especially in the setting without lexical overlap initialization ($*\textsc{-rand}$). This indicates that by factorizing the information encoded in the original embedding matrix we are able to extract relevant information for unseen languages. Combining matrix factorization with lexical overlap initialization (\textsc{MF$_*^*\text{-lex}$}), zero-shot performance improves further for unseen languages with covered scripts. This suggests that the two methods complement each other. For 6/9 of these languages, we find that encoding the embeddings in multiple up-projections (\textsc{MF$_*^{10}\text{-}*$}) achieves the peak score. This in turn verifies 
that grouping  similar tokens improves the robustness of token representations \cite{ChungGTR20}. For unseen languages with covered scripts, this model variant also outperforms \madxtwo on average. 

For languages with unseen scripts we find that MF has smaller impact. While the encoded information supports languages similar to those seen by the model in pretraining, languages with unseen scripts are too distant to benefit from this latent multilingual knowledge. Surprisingly, lexical overlap is helpful for languages with unseen scripts.

Overall, we observe that both \textsc{MF$_{\textsc{KMeans}}^{10}\text{-}*$} and \textsc{MF$_{\textsc{Neural}}^{10}\text{-}*$} perform well for most languages, where the {\footnotesize\textsc{KMeans}} variant performs better for NER and the {\footnotesize\textsc{Neural}} variant performs better for UD.

\begin{table*}
\centering
\footnotesize
\resizebox{\textwidth}{!}{
\begin{tabular}{lrrr|rrrrrrrrr|rrrrr}
\toprule
    & \multicolumn{3}{c}{Seen Languages} & \multicolumn{9}{c}{Unseen Languages but Covered Scripts} & \multicolumn{5}{c}{New Scripts} \\  

	 & 	ka & ur & hi  & cdo & mi & ilo & gn & xmf & sd & myv & bh & wo & am & bo & km & dv & si  \\
\midrule

\mflexkmeans 100 & 66.3 & 59.5 & 61.1  & 48.9 & 47.5 & 46.4 & 53.6 & 56.7 & 46.2 & 58.4 & 61.0 & 57.9 & 45.5 & 51.6 & 64.2 & 44.9 & 46.2  \\
\mflexkmeans 300 & 65.1 & 46.8 & 56.8 & 26.6 & 5.5 & 53.7 & 10.0 & 52.2 & 43.5 & 37.4 & 53.2 & 3.7 & 21.5 & 7.4 & 63.9 & 38.8 & 51.2 \\
\bottomrule
\end{tabular}
}
\caption{Mean $F_1$ test results averaged over 5 runs and  the 4 high-resource source languages English, Chinese, Japanese, and Arabic. The first row presents results with 100-, the second row 300-dimensional embeddings.}
\label{tab:parameter_eff}
\end{table*}

\section{Further Analysis}

\subsection{Lexically Overlapping (Sub)Words}
We perform a quantitative analysis of lexically overlapping tokens, i.e. that occur both in mBERT's and monolingual in-language vocabularies; see Table \ref{tab:lex-overlap-groupings}.
For languages with scripts not covered by the mBERT tokenizer, most lexically-overlapping tokens are single characters of a non-Latin script.

We further present the top 5 longest lexically overlapping (sub) words for four languages with scripts covered during pretraining (Min Dong, Maori, Ilokano, and Guarani) and four languages with unseen scripts (Tibetan, Khmer, Divehi, Sinhala) in Table \ref{table:top_lex_overlap}. We observe that frequent lexically overlapping tokens are named entities in the Latin script,  
indicating that NER may not be the best evaluation task to objectively assess generalization performance to such languages. If the same named entities also occur in the training data of higher-resource languages, the models will be more successful at identifying them in the unseen language, which belies a lack of deeper understanding of the low-resource language. This might also explain  why greater performance gains were achieved for NER  
than for DP.

 \begin{figure}[!t]
 
        \centering
        \includegraphics[width=\linewidth]{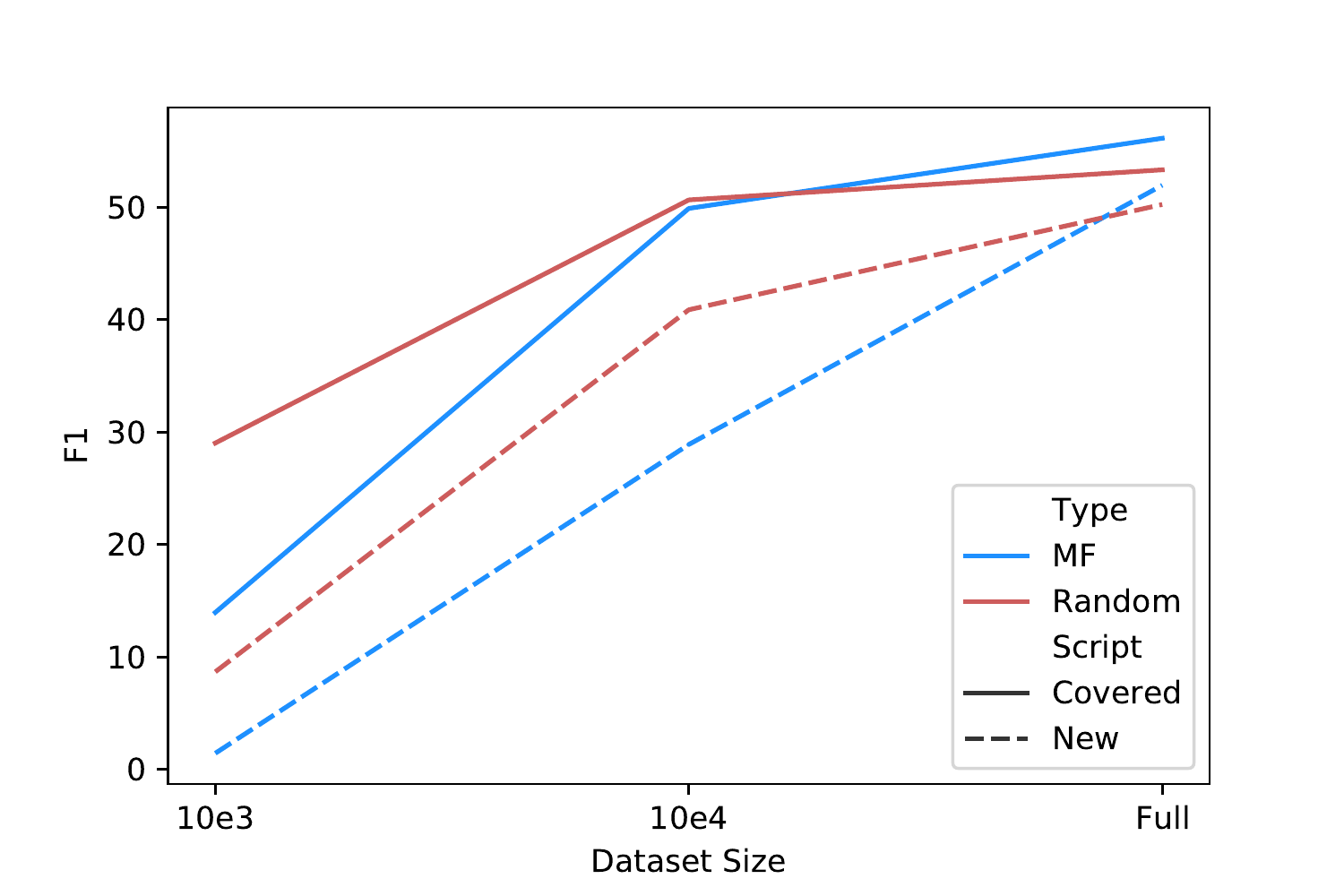}

    \vspace{-1mm}
    \caption{Sample efficiency. For "MF" we leverage the \mflexkmeans setting.}
\label{fig:sample_eff}
\vspace{-1.1em}
\end{figure}

\subsection{Sample Efficiency}

We have further analyzed the sample efficiency of our approaches. We find that \textsc{EL-lex} is slightly more sample-efficient than \textsc{MF$_{\text{KMeans}}^{10}\text{-lex}$}, where the latter outperforms the former with more data available. In Figure \ref{fig:sample_eff}  we plot the zero-shot transfer performance where the adapters and embeddings were pretrained on different amounts of data. We further find that lower-dimensional embeddings tend to outperform higher-dimensional embeddings for the majority of languages. In Table \ref{tab:parameter_eff} we compare 300 with 100 dimensional embeddings for the \textsc{MF$_{\text{KMeans}}^{10}\text{-lex}$} approach.

\subsection{Script Clusters}

We analyzed the KMeans clusters based on the tokens that consist of characters of a certain script in Figure~\ref{fig:kmeans_clusters} of the Appendix. We find  distinct script-based groups; For instance, 5 clusters consist primarily of Latin-script tokens\footnote{The majority of tokens of mBERT use the Latin script.}, two clusters predominantly consist of Chinese, and a few Korean tokens. Interestingly, 2 clusters consisted of Cyrilic, and Arabic scripts as well as scripts used predominantly in India, varying slightly in their distribution. Lastly, one cluster included tokens of all except the Latin script.

\section{Conclusion}
We have systematically evaluated strategies for model adaptation to unseen languages with seen and unseen scripts. We have assessed the importance of the information stored within the original embedding matrix by  means of leveraging lexically overlapping tokens, and extracting latent semantic concepts. For the latter, we have proposed a new method of encoding the embedding matrix into lower- dimensional embeddings and up-projections. We have demonstrated that our methods outperform previous approaches on NER and dependency parsing for both resource-rich and resource-poor scenarios, reducing the gap between random and lexical overlap initialisation, and enabling more effective model adaptation to unseen scripts.

\section*{Acknowledgments}

Jonas Pfeiffer is supported by the LOEWE initiative (Hesse, Germany) within the emergenCITY center.  
The work of Ivan Vuli\'{c} is supported by the ERC Grant LEXICAL (no. 648909) and the ERC PoC Grant MultiConvAI (no. 957356).

We thank Laura Rimell, Nils Reimers, Michael Bugert and the anonymous reviewers for insightful feedback and suggestions on a draft of this paper.

\bibliographystyle{acl_natbib}
\bibliography{anthology,acl2021}


\appendix

\section{Appendices}

\subsection{Full Model Transfer}
\label{sec:ap_full_model_transfer}
For comparability with previous work, which generally fine-tunes the entire model \cite{ChauLS20Parsing,Wang20ExtendmBERT}, we follow \citet{Artetxe2020cross-lingual} and learn a new embedding matrix for the target language while freezing the pretrained transformer weights. For training on the target task, we fine-tune the transformer weights of the pretrained model while keeping the original embedding layer frozen. For zero-shot transfer, we replace the existing with the new embedding matrix trained on the target language.

\subsection{Results: Named Entity Recognition}

We present non-aggregated NER transfer performance when transferring from English, Chinese, Japanese, and Arabic in Tables \ref{tab:english_zeroshot}, \ref{tab:chinese_zeroshot}, \ref{tab:Japanese_zeroshot}, and \ref{tab:appendix_zeroshot_ner}  
respectively.  \textit{12Ad} indicates whether (\cmark) or not (\xmark) an adapter is placed in the 12th transformer layer.  We additionally present the results for full model fine-tuning (\textsc{FMT}-$*$).  

\subsection{Results: Dependency Parsing}

We present non-aggregated DP transfer performance when transferring from English, Chinese, Japanese, and Arabic in Tables \ref{tab:mainresults_UD_english}, \ref{tab:mainresults_UD_chinese}, \ref{tab:mainresults_UD_japanese}, and \ref{tab:mainresults_UD_arabic} respectively. 
 
\subsection{Script Clusters}

\begin{figure*}[htp]
    \centering
 
        \centering
        \includegraphics[width=2.09\columnwidth]{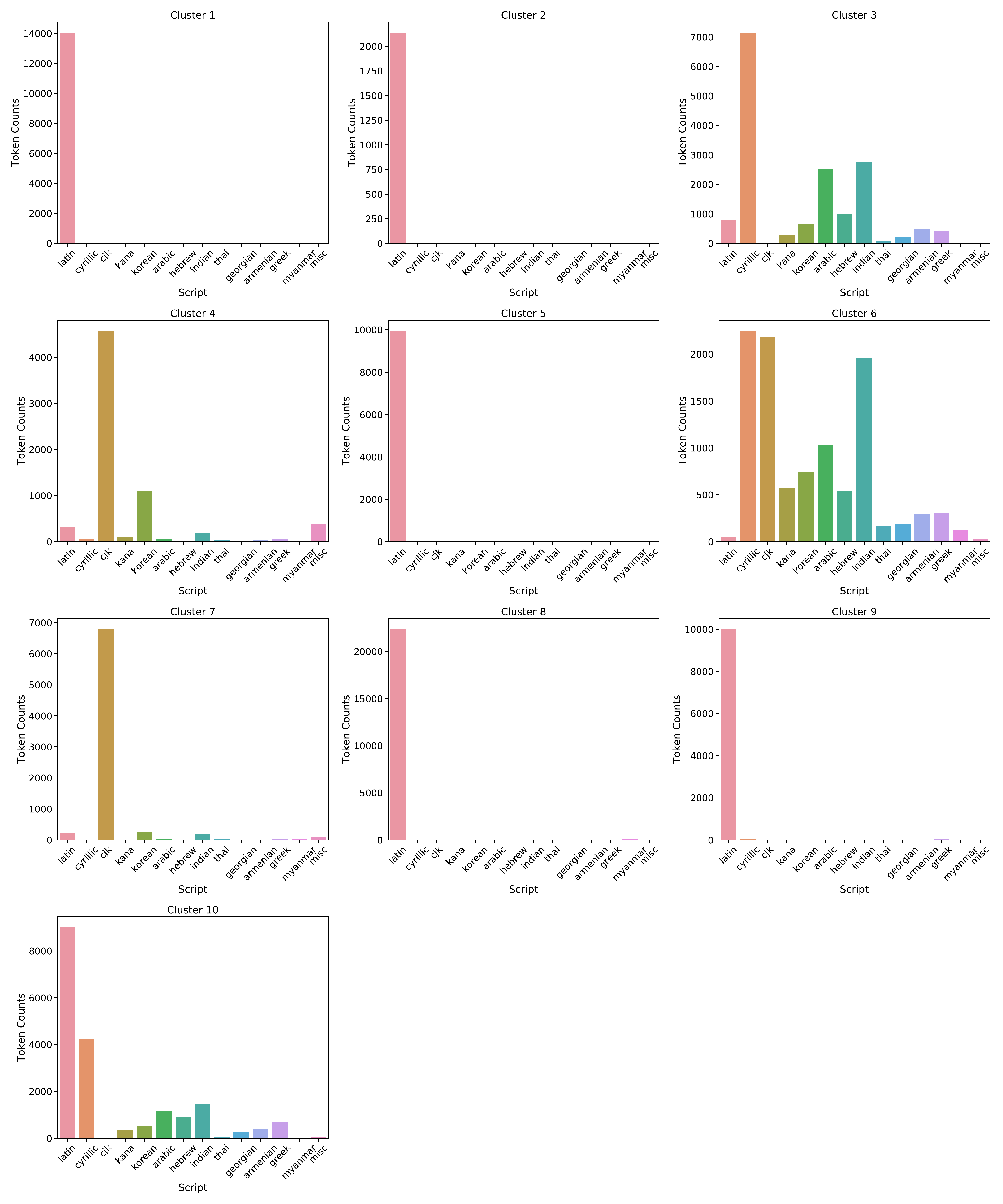}

    \vspace{-1mm}
    \caption{Number of tokens of each script grouped into the 10 KMeans clusters. We follow \cite{acs:2019} in splitting the utf-8 tokens into scripts.  }
    \vspace{-1mm}
\label{fig:kmeans_clusters}
\end{figure*}

We present the groups of scripts within the 10 KMeans clusters in Figure~\ref{fig:kmeans_clusters}. We follow \citet{acs:2019} in grouping the scripts into languages. 

  \subsection{Language information}

Table~\ref{table:list_of_languages} lists all 104 languages and corresponding scripts which mBERT was pretrained on. 
 
\subsection{Hardware Setup}
All experiments were conducted on a single NVIDIA V100 GPU with 32 Gb of VRAM.

\begin{table*}

\centering
\footnotesize
\resizebox{\textwidth}{!}{
\begin{tabular}{lcrrrr|rrrrrrrrrr|rrrrrr}
\toprule
 &\multirow{2}{*}{\rotatebox[origin=c]{90}{\begin{tabular}[c]{@{}c@{}}12Ad\end{tabular}}}   & \multicolumn{4}{c}{Seen Languages} & \multicolumn{10}{c}{Unseen Languages but Covered Scripts} & \multicolumn{6}{c}{New Scripts} \\   

 & 	 & 	ka & ur & hi & \textbf{Avg} & cdo & mi & ilo & gn & xmf & sd & myv & bh & wo & \textbf{Avg} & am & bo & km & dv & si & \textbf{Avg}  \\
\midrule
mBERT & & 64.3 & 34.3 & 65.3 & 54.6 & 17.3 & 22.2 & 62.7 & 46.7 & 34.4 & 12.1 & 47.2 & 48.8 & 26.7 & 35.4 & 0.0 & 11.4 & 4.9 & 0.3 & 0.3 & 3.4 \\
\textsc{MAD-X} &  \cmark&  64.8 & 44.5 & 60.7 & 56.7 & 25.1 & 38.4 & 61.2 & 52.7 & 46.0 & 25.3 & 47.7 & 52.1 & 47.1 & 43.9 & 14.5 & 20.4 & 14.5 & 13.9 & 16.7 & 16.0 \\
\textsc{MAD-X 2.0} & \xmark &  67.4 & 54.4 & 64.3 & 62.1 & 24.5 & 51.8 & 79.2 & 54.4 & 49.0 & 30.9 & 58.1 & 55.6 & 56.8 & 51.1 & 8.0 & 16.6 & 18.4 & 23.0 & 22.0 & 17.6 \\
\midrule
\textsc{FMT-RandInit}  &  & 65.1 & 48.9 & 63.3 & 59.1 & 26.9 & 12.6 & 67.1 & 20.9 & 54.6 & 37.5 & 8.0 & 47.1 & 10.9 & 31.7 & 28.6 & 29.3 & 51.6 & 27.0 & 50.0 & 37.3	 \\
\elrand	 &\cmark  & 66.6 & 35.0 & 62.9 & 54.8 & 33.5 & 11.0 & 58.8 & 49.0 & 47.8 & 37.5 & 20.0 & 53.6 & 4.1 & 35.0 & 36.8 & 40.2 & 57.6 & 35.5 & 48.4 & 43.7	 \\
\elrand	 & \xmark & 70.1 & 39.2 & 67.5 & 58.9 & 34.0 & 16.4 & 66.6 & 53.0 & 57.9 & 39.7 & 35.1 & 56.4 & 4.9 & 40.4 & 46.4 & 50.8 & 61.4 & 38.9 & 52.6 & 50.0	 \\
\mfrand &\cmark &  62.1 & 46.2 & 61.2 & 56.5 & 29.5 & 28.1 & 51.4 & 41.6 & 47.3 & 40.8 & 38.5 & 33.8 & 29.0 & 37.8 & 39.6 & 39.3 & 54.7 & 35.4 & 46.6 & 43.1 \\
\mfrand & \xmark&  67.2 & 51.5 & 63.2 & 60.6 & 33.3 & 33.7 & 58.0 & 48.6 & 56.1 & 44.3 & 55.2 & 43.2 & 41.8 & 46.0 & 43.3 & 42.1 & 63.4 & 41.6 & 54.7 & 49.0 \\ 
\mfrandkmeans &\cmark & 63.2 & 52.0 & 61.4 & 58.9 & 28.4 & 7.2 & 52.8 & 45.6 & 51.3 & 15.3 & 30.0 & 47.8 & 32.8 & 34.6 & 37.4 & 34.1 & 55.1 & 33.7 & 44.4 & 40.9\\
\mfrandkmeans &\xmark & 67.3 & 55.5 & 63.9 & 62.3 & 35.1 & 6.3 & 65.0 & 57.3 & 59.1 & 20.4 & 44.8 & 55.2 & 42.6 & 42.9 & 46.8 & 41.8 & 60.1 & 37.5 & 51.0 & 47.4\\
\mfrandnn &\cmark &     63.9 & 51.7 & 62.6 & 59.4 & 14.0 & 4.1 & 57.0 & 49.4 & 48.2 & 35.6 & 23.6 & 52.4 & 34.8 & 35.5 & 31.6 & 32.0 & 58.3 & 17.2 & 35.6 & 34.9\\
\mfrandnn &\xmark &     68.4 & 56.3 & 62.8 & 62.5 & 16.0 & 4.7 & 65.9 & 56.4 & 57.6 & 42.2 & 39.1 & 58.5 & 44.8 & 42.8 & 37.3 & 31.0 & 61.7 & 18.9 & 42.2 & 38.2\\
\cdashline{1-21} \vspace{-3mm} \\
\textsc{FMT-LexInit}  &  & 	67.6 & 51.8 & 63.1 & 60.9 & 42.2 & 31.3 & 72.4 & 47.6 & 56.7 & 35.6 & 53.1 & 53.4 & 45.6 & 48.7 & 34.9 & 25.9 & 49.9 & 43.9 & 55.4 & 42.0 \\
\ellex & \cmark & 70.1 & 51.4 & 65.2 & 62.2 & 42.5 & 37.9 & 56.8 & 50.8 & 52.5 & 41.1 & 38.8 & 56.1 & 54.2 & 47.9 & 40.0 & 41.9 & 60.8 & 43.3 & 52.6 & 47.7 \\
\ellex & \xmark & 73.1 & 56.0 & 69.2 & 66.1 & 42.9 & 34.3 & 66.1 & 54.9 & 56.9 & 44.0 & 50.8 & 59.4 & 61.4 & 52.3 & 46.5 & 46.6 & 60.0 & 48.3 & 60.3 & 52.3 \\
\mflex &\cmark  & 66.4 & 41.1 & 62.0 & 56.5 & 41.3 & 37.2 & 57.1 & 46.5 & 47.0 & 35.7 & 29.5 & 50.6 & 36.5 & 42.4 & 36.4 & 40.0 & 59.1 & 35.6 & 44.1 & 43.0\\
\mflex & \xmark & 71.0 & 51.7 & 65.6 & 62.8 & 43.3 & 55.6 & 64.9 & 57.5 & 53.2 & 42.1 & 47.5 & 57.4 & 45.9 & 51.9 & 43.1 & 41.8 & 63.1 & 44.0 & 53.6 & 49.1\\ 
\mflexkmeans &\cmark & 68.0 & 56.4 & 63.7 & 62.7 & 38.1 & 45.5 & 60.2 & 47.8 & 46.5 & 39.8 & 33.5 & 52.2 & 50.1 & 46.0 & 42.2 & 36.8 & 61.6 & 39.1 & 50.5 & 46.0 \\
\mflexkmeans &\xmark & 71.0 & 58.0 & 65.8 & 64.9 & 44.7 & 65.7 & 67.1 & 55.5 & 55.2 & 43.9 & 41.0 & 57.3 & 59.2 & 54.4 & 49.0 & 36.3 & 65.3 & 45.5 & 55.3 & 50.3 \\
\mflexnn &\cmark &     65.7 & 53.6 & 60.2 & 59.8 & 43.7 & 51.1 & 52.7 & 51.0 & 46.7 & 41.1 & 46.6 & 55.9 & 48.8 & 48.6 & 43.8 & 37.3 & 58.9 & 39.1 & 41.1 & 44.0  \\
\mflexnn &\xmark &     70.6 & 60.5 & 63.9 & 65.0 & 45.3 & 65.0 & 61.9 & 53.6 & 53.0 & 43.2 & 56.9 & 60.2 & 60.2 & 55.5 & 53.5 & 46.1 & 61.1 & 44.5 & 43.2 & 49.7  \\
\bottomrule
\end{tabular}
}
\caption{Mean $F_1$ NER test results averaged over 5 runs transferring from high resource language  \textbf{English} to the low resource languages. \textit{12Ad} indicates whether (\cmark) or not (\xmark) an adapter is placed in the 12th transformer layer.  The top group (first three rows) includes models which leverage the original tokenizer which is not specialized for the target language. The second group (last 18 rows) include models with new tokenizers. Here we separate models with randomly initialized embeddings ($*$-\textsc{RandInit}) from models with lexical initialization ($*$-\textsc{LexInit}) by the dashed line. We additionally present the results for full model fine-tuning (\textsc{FMT}-$*$). 
}
\label{tab:english_zeroshot}
\end{table*}

\vspace{1.0em}
 
\begin{table*}[t]

\centering
\footnotesize
\resizebox{\textwidth}{!}{
\begin{tabular}{lcrrrr|rrrrrrrrrr|rrrrrr}
\toprule
 &\multirow{2}{*}{\rotatebox[origin=c]{90}{\begin{tabular}[c]{@{}c@{}}12Ad\end{tabular}}}   & \multicolumn{4}{c}{Seen Languages} & \multicolumn{10}{c}{Unseen Languages but Covered Scripts} & \multicolumn{6}{c}{New Scripts} \\  

 & 	 & 	ka & ur & hi & \textbf{Avg} & cdo & mi & ilo & gn & xmf & sd & myv & bh & wo & \textbf{Avg} & am & bo & km & dv & si & \textbf{Avg}  \\
\midrule
mBERT & & 66.0 & 33.3 & 60.2 & 53.2 & 21.2 & 42.3 & 42.6 & 59.5 & 49.1 & 14.2 & 29.4 & 60.1 & 25.1 & 38.2 & 0.2 & 13.1 & 21.8 & 1.4 & 2.0 & 7.7 \\
\textsc{MAD-X} &  \cmark&  62.9 & 50.8 & 54.1 & 55.9 & 32.2 & 42.5 & 40.6 & 56.4 & 57.3 & 44.4 & 68.6 & 63.1 & 55.1 & 51.1 & 14.6 & 24.4 & 24.1 & 19.8 & 18.9 & 20.4 \\
\textsc{MAD-X 2.0} & \xmark &  63.9 & 52.4 & 55.7 & 57.3 & 37.0 & 42.1 & 64.7 & 57.9 & 54.9 & 42.8 & 72.0 & 63.8 & 60.1 & 55.1 & 13.5 & 29.0 & 23.0 & 21.5 & 20.2 & 21.4 \\
\midrule
\textsc{FMT-RandInit}  &  & 64.6 & 37.2 & 56.5 & 52.8 & 36.7 & 4.1 & 48.1 & 15.7 & 57.8 & 45.8 & 7.0 & 48.9 & 15.8 & 31.1 & 26.4 & 28.8 & 53.0 & 31.1 & 41.6 & 36.2	 \\
\elrand	 &\cmark  & 62.8 & 43.7 & 57.2 & 54.6 & 41.4 & 6.8 & 51.7 & 45.5 & 57.4 & 48.1 & 38.7 & 58.4 & 13.4 & 40.2 & 41.8 & 49.8 & 51.6 & 30.1 & 54.3 & 45.5	 \\
\elrand	 & \xmark & 63.0 & 48.3 & 59.3 & 56.9 & 44.8 & 4.4 & 54.3 & 47.5 & 60.1 & 47.0 & 46.5 & 58.6 & 8.7 & 41.3 & 43.5 & 51.8 & 53.6 & 29.7 & 51.3 & 46.0	 \\
\mfrand &\cmark &  59.3 & 38.1 & 50.3 & 49.2 & 38.2 & 10.0 & 34.7 & 44.4 & 54.1 & 46.5 & 59.8 & 43.0 & 47.8 & 42.1 & 41.6 & 49.1 & 59.7 & 42.3 & 53.6 & 49.2 \\
\mfrand & \xmark&  59.5 & 42.8 & 52.9 & 51.7 & 44.2 & 10.2 & 35.9 & 44.3 & 54.2 & 44.8 & 64.2 & 43.4 & 47.4 & 43.2 & 43.9 & 48.6 & 55.9 & 45.3 & 52.4 & 49.2 \\ 
\mfrandkmeans &\cmark & 60.0 & 40.7 & 55.5 & 52.1 & 38.6 & 1.0 & 43.2 & 44.4 & 64.3 & 28.6 & 55.0 & 61.0 & 50.9 & 43.0 & 39.3 & 23.5 & 63.7 & 34.6 & 45.0 & 41.2 \\
\mfrandkmeans &\xmark & 59.6 & 43.9 & 56.8 & 53.4 & 48.1 & 3.1 & 47.6 & 50.3 & 63.1 & 33.0 & 60.6 & 59.7 & 51.9 & 46.4 & 44.9 & 25.3 & 62.1 & 37.5 & 45.7 & 43.1 \\
\mfrandnn &\cmark &     62.5 & 44.1 & 54.1 & 53.5 & 21.3 & 2.2 & 44.7 & 46.1 & 58.4 & 50.8 & 39.7 & 63.9 & 47.6 & 41.6 & 37.5 & 38.7 & 65.4 & 18.3 & 50.8 & 42.1 \\
\mfrandnn &\xmark &     62.1 & 50.0 & 56.3 & 56.1 & 20.7 & 4.1 & 49.7 & 51.4 & 57.9 & 52.5 & 48.2 & 65.6 & 48.3 & 44.3 & 38.3 & 42.9 & 64.3 & 18.1 & 52.5 & 43.2 \\
\cdashline{1-21} \vspace{-3mm} \\
\textsc{FMT-LexInit}  &  & 	67.1 & 36.9 & 60.2 & 54.7 & 59.6 & 15.6 & 49.3 & 55.4 & 57.3 & 45.5 & 52.3 & 64.6 & 50.4 & 50.0 & 29.3 & 31.3 & 58.2 & 45.2 & 48.6 & 42.5 \\
\ellex & \cmark & 68.9 & 49.7 & 61.7 & 60.1 & 51.4 & 26.6 & 54.3 & 53.5 & 54.6 & 48.4 & 59.7 & 62.2 & 59.3 & 52.2 & 43.9 & 47.1 & 60.2 & 48.3 & 55.7 & 51.0 \\
\ellex & \xmark & 68.0 & 53.1 & 64.0 & 61.7 & 54.7 & 24.4 & 54.4 & 55.3 & 52.4 & 48.4 & 60.7 & 59.6 & 58.1 & 52.0 & 48.0 & 53.5 & 60.7 & 50.3 & 54.8 & 53.5 \\
\mflex &\cmark  & 61.8 & 42.8 & 56.4 & 53.7 & 51.2 & 16.3 & 47.2 & 49.7 & 52.6 & 47.2 & 56.2 & 60.0 & 53.7 & 48.2 & 41.5 & 39.4 & 61.4 & 41.7 & 44.8 & 45.8\\
\mflex & \xmark & 61.9 & 47.8 & 57.7 & 55.8 & 48.2 & 18.8 & 51.5 & 55.8 & 54.9 & 46.7 & 61.3 & 60.4 & 52.6 & 50.0 & 42.9 & 41.9 & 58.3 & 44.8 & 49.6 & 47.5\\ 
\mflexkmeans &\cmark & 62.9 & 47.2 & 58.3 & 56.1 & 49.2 & 31.8 & 53.8 & 53.2 & 51.1 & 50.1 & 56.0 & 61.6 & 58.5 & 51.7 & 42.7 & 43.6 & 62.4 & 46.9 & 50.0 & 49.1 \\
\mflexkmeans &\xmark & 63.3 & 50.9 & 59.8 & 58.0 & 51.8 & 29.6 & 52.5 & 55.3 & 51.5 & 50.7 & 59.9 & 65.1 & 56.0 & 52.5 & 44.9 & 44.4 & 61.0 & 49.9 & 48.0 & 49.6 \\
\mflexnn &\cmark &     62.6 & 55.1 & 55.6 & 57.8 & 52.9 & 32.3 & 40.0 & 54.4 & 59.4 & 53.4 & 61.1 & 60.8 & 67.4 & 53.5 & 44.3 & 50.1 & 65.7 & 40.9 & 53.4 & 50.9  \\
\mflexnn &\xmark &     63.7 & 57.1 & 58.4 & 59.7 & 54.1 & 34.2 & 42.0 & 54.6 & 59.0 & 49.6 & 63.4 & 62.9 & 66.9 & 54.1 & 41.5 & 53.8 & 62.3 & 45.4 & 49.6 & 50.5  \\
\bottomrule
\end{tabular}
}
\caption{ Mean $F_1$ NER test results averaged over 5 runs transferring from high resource language  \textbf{Chinese} to the low resource languages. \textit{12Ad} indicates whether (\cmark) or not (\xmark) an adapter is placed in the 12th transformer layer.  The top group (first three rows) includes models which leverage the original tokenizer which is not specialized for the target language. The second group (last 18 rows) include models with new tokenizers. Here we separate models with randomly initialized embeddings ($*$-\textsc{RandInit}) from models with lexical initialization ($*$-\textsc{LexInit}) by the dashed line. We additionally present the results for full model fine-tuning (\textsc{FMT}-$*$). 
}
\label{tab:chinese_zeroshot}
\end{table*}

 \vspace{1.0em}
 
\begin{table*}[t]
\centering
\footnotesize
\resizebox{\textwidth}{!}{
\begin{tabular}{lcrrrr|rrrrrrrrrr|rrrrrr}
\toprule
 &\multirow{2}{*}{\rotatebox[origin=c]{90}{\begin{tabular}[c]{@{}c@{}}12Ad\end{tabular}}}   & \multicolumn{4}{c}{Seen Languages} & \multicolumn{10}{c}{Unseen Languages but Covered Scripts} & \multicolumn{6}{c}{New Scripts} \\  

 & 	 & 	ka & ur & hi & \textbf{Avg} & cdo & mi & ilo & gn & xmf & sd & myv & bh & wo & \textbf{Avg} & am & bo & km & dv & si & \textbf{Avg}  \\
\midrule
mBERT & &  64.2 & 35.5 & 59.8 & 53.2 & 17.7 & 42.6 & 39.5 & 50.2 & 46.5 & 14.2 & 23.9 & 56.0 & 21.6 & 34.7 & 3.4 & 23.2 & 6.2 & 1.4 & 7.6 & 8.3\\
\textsc{MAD-X} &  \cmark & 62.7 & 49.8 & 55.9 & 56.1 & 29.7 & 43.8 & 28.3 & 48.0 & 53.2 & 34.2 & 58.0 & 54.7 & 52.9 & 44.7 & 6.0 & 31.4 & 22.4 & 16.7 & 17.2 & 18.7    \\
\textsc{MAD-X 2.0} & \xmark &  64.7 & 52.5 & 57.8 & 58.3 & 31.8 & 46.6 & 43.5 & 54.6 & 51.7 & 40.0 & 60.6 & 59.9 & 50.8 & 48.8 & 7.2 & 31.2 & 22.7 & 24.4 & 22.1 & 21.5 \\
\midrule
\textsc{FMT-RandInit}  &  & 64.2 & 36.4 & 60.2 & 53.6 & 33.3 & 3.8 & 49.7 & 15.5 & 56.1 & 47.2 & 11.7 & 54.6 & 12.7 & 31.6 & 27.5 & 33.0 & 58.5 & 30.4 & 52.5 & 40.4	 \\
\elrand	 &\cmark  & 63.0 & 37.1 & 58.2 & 52.8 & 35.8 & 3.9 & 47.3 & 43.7 & 50.1 & 41.6 & 22.5 & 54.7 & 5.7 & 33.9 & 34.9 & 50.6 & 52.7 & 32.3 & 44.6 & 43.0	 \\
\elrand	 & \xmark & 65.6 & 43.1 & 60.7 & 56.5 & 37.4 & 2.6 & 52.6 & 47.4 & 58.1 & 46.7 & 31.2 & 56.8 & 4.3 & 37.5 & 40.3 & 53.5 & 60.7 & 32.3 & 52.8 & 47.9\\
\mfrand &\cmark &  60.8 & 43.1 & 52.7 & 52.2 & 36.7 & 9.1 & 32.4 & 42.0 & 55.2 & 42.1 & 57.1 & 43.1 & 46.7 & 40.5 & 37.2 & 48.3 & 56.8 & 37.9 & 50.6 & 46.2 \\
\mfrand & \xmark&  63.5 & 44.6 & 55.7 & 54.6 & 43.3 & 12.2 & 28.9 & 44.9 & 57.7 & 48.5 & 62.4 & 46.0 & 55.9 & 44.4 & 41.4 & 46.8 & 57.6 & 40.1 & 52.8 & 47.7 \\ 
\mfrandkmeans &\cmark & 58.8 & 36.5 & 56.5 & 50.6 & 34.5 & 0.9 & 24.0 & 45.7 & 57.8 & 20.6 & 27.3 & 52.4 & 42.2 & 33.9 & 33.5 & 39.9 & 61.9 & 34.7 & 40.7 & 42.1 \\
\mfrandkmeans &\xmark & 64.4 & 39.0 & 58.4 & 53.9 & 40.3 & 1.3 & 31.3 & 48.8 & 57.0 & 26.5 & 46.1 & 55.5 & 43.8 & 39.0 & 36.2 & 40.7 & 63.6 & 37.0 & 47.2 & 44.9 \\
\mfrandnn &\cmark &     59.4 & 37.4 & 55.4 & 50.7 & 22.6 & 2.6 & 29.3 & 44.8 & 52.1 & 45.5 & 28.3 & 58.5 & 41.3 & 36.1 & 31.0 & 42.4 & 63.9 & 17.0 & 45.5 & 40.0 \\
\mfrandnn &\xmark &     62.8 & 44.4 & 56.7 & 54.6 & 19.5 & 3.7 & 32.6 & 46.0 & 57.3 & 44.5 & 45.4 & 62.3 & 42.9 & 39.4 & 33.8 & 40.1 & 63.7 & 18.2 & 44.5 & 40.0 \\
\cdashline{1-21} \vspace{-3mm} \\
\textsc{FMT-LexInit}  &  & 	66.1 & 40.4 & 62.0 & 56.2 & 52.4 & 13.7 & 45.5 & 52.8 & 56.7 & 43.0 & 61.2 & 64.0 & 50.2 & 48.9 & 31.5 & 30.0 & 60.7 & 40.2 & 53.9 & 43.3 \\
\ellex & \cmark & 60.3 & 51.7 & 61.8 & 57.9 & 42.3 & 20.1 & 55.3 & 47.5 & 50.8 & 42.7 & 60.9 & 59.8 & 55.2 & 48.3 & 39.8 & 44.8 & 60.2 & 39.9 & 52.7 & 47.5 \\
\ellex & \xmark & 68.8 & 52.5 & 62.9 & 61.4 & 46.6 & 22.6 & 51.3 & 50.4 & 55.0 & 44.7 & 60.1 & 63.4 & 56.0 & 50.0 & 45.0 & 47.3 & 61.7 & 45.4 & 57.7 & 51.4 \\
\mflex &\cmark  & 62.7 & 47.4 & 58.6 & 56.2 & 47.0 & 22.1 & 38.0 & 49.6 & 51.6 & 42.0 & 57.1 & 59.2 & 55.2 & 46.9 & 34.9 & 41.4 & 63.3 & 41.1 & 43.0 & 44.8\\
\mflex & \xmark & 64.6 & 52.1 & 59.4 & 58.7 & 48.8 & 26.7 & 43.0 & 51.0 & 55.1 & 44.6 & 58.3 & 63.1 & 57.3 & 49.8 & 32.2 & 42.0 & 63.8 & 43.1 & 50.4 & 46.3\\ 
\mflexkmeans &\cmark & 62.2 & 46.9 & 57.4 & 55.5 & 43.2 & 33.6 & 31.8 & 46.1 & 53.5 & 42.3 & 35.8 & 60.2 & 55.8 & 44.7 & 37.8 & 41.9 & 60.9 & 44.7 & 50.5 & 47.2 \\
\mflexkmeans &\xmark & 67.0 & 51.5 & 59.4 & 59.3 & 51.2 & 45.8 & 35.6 & 49.9 & 55.0 & 45.8 & 50.5 & 64.3 & 56.7 & 50.5 & 42.6 & 38.7 & 63.6 & 44.3 & 51.2 & 48.1 \\
\mflexnn &\cmark &     62.1 & 49.2 & 56.4 & 55.9 & 45.1 & 20.5 & 29.2 & 47.1 & 51.3 & 45.9 & 48.2 & 62.3 & 62.2 & 45.8 & 33.6 & 50.4 & 67.1 & 40.0 & 45.9 & 47.4  \\
\mflexnn &\xmark &     66.2 & 52.4 & 58.5 & 59.0 & 48.4 & 36.0 & 29.6 & 51.5 & 54.9 & 46.8 & 56.3 & 63.6 & 65.9 & 50.3 & 38.1 & 54.7 & 64.6 & 43.6 & 46.8 & 49.6  \\
\bottomrule
\end{tabular}
}
\caption{Mean $F_1$ NER test results averaged over 5 runs transferring from high resource language  \textbf{Japanese} to the low resource languages. \textit{12Ad} indicates whether (\cmark) or not (\xmark) an adapter is placed in the 12th transformer layer.  The top group (first three rows) includes models which leverage the original tokenizer which is not specialized for the target language. The second group (last 18 rows) include models with new tokenizers. Here we separate models with randomly initialized embeddings ($*$-\textsc{RandInit}) from models with lexical initialization ($*$-\textsc{LexInit}) by the dashed line. We additionally present the results for full model fine-tuning (\textsc{FMT}-$*$). 
}
\label{tab:Japanese_zeroshot}
 
 \end{table*}
 
\begin{table*}
 
\centering
\footnotesize
\resizebox{\textwidth }{!}{
\begin{tabular}{lcrrrr|rrrrrrrrrr|rrrrrr}
\toprule
 &\multirow{2}{*}{\rotatebox[origin=c]{90}{\begin{tabular}[c]{@{}c@{}}12Ad\end{tabular}}}   & \multicolumn{4}{c}{Seen Languages} & \multicolumn{10}{c}{Unseen Languages but Covered Scripts} & \multicolumn{6}{c}{New Scripts} \\  

 & 	 & 	ka & ur & hi & \textbf{Avg} & cdo & mi & ilo & gn & xmf & sd & myv & bh & wo & \textbf{Avg} & am & bo & km & dv & si & \textbf{Avg}  \\
\midrule
mBERT & & 64.4 & 46.2 & 70.0 & 60.2 & 17.0 & 25.8 & 36.7 & 58.4 & 43.6 & 10.3 & 22.3 & 53.2 & 23.6 & 32.3 & 0.0 & 21.9 & 9.8 & 1.9 & 0.3 & 6.8 \\
\textsc{MAD-X} &  \cmark&  65.3 & 60.6 & 64.1 & 63.4 & 29.5 & 57.8 & 52.0 & 53.5 & 51.2 & 32.3 & 57.0 & 57.9 & 43.2 & 48.3 & 7.9 & 22.9 & 9.6 & 16.8 & 14.4 & 14.3 \\
\textsc{MAD-X 2.0} & \xmark &  66.5 & 61.9 & 66.1 & 64.9 & 29.5 & 61.8 & 68.7 & 58.4 & 54.8 & 34.8 & 61.3 & 59.8 & 54.5 & 53.7 & 14.1 & 22.0 & 8.5 & 20.0 & 10.4 & 15.0 \\
\midrule
\textsc{FMT-RandInit}  &  & 64.3 & 44.0 & 66.9 & 58.4 & 37.2 & 2.0 & 59.6 & 20.0 & 64.9 & 41.1 & 9.1 & 55.8 & 4.9 & 32.7 & 30.2 & 38.3 & 56.7 & 31.7 & 54.7 & 42.3	 \\
\elrand	 &\cmark  & 61.8 & 59.1 & 64.9 & 61.9 & 31.8 & 6.7 & 52.7 & 43.3 & 55.4 & 38.0 & 29.4 & 48.6 & 3.0 & 34.3 & 36.9 & 59.8 & 61.4 & 27.7 & 44.3 & 46.0	 \\
\elrand	 & \xmark & 64.4 & 60.7 & 67.8 & 64.3 & 35.9 & 5.9 & 58.1 & 45.9 & 61.5 & 43.6 & 27.8 & 50.4 & 2.8 & 36.9 & 41.4 & 59.4 & 62.2 & 30.1 & 49.2 & 48.5	 \\
\mfrand &\cmark & 61.8 & 52.9 & 57.8 & 57.5 & 36.4 & 18.5 & 46.8 & 49.3 & 59.0 & 39.0 & 57.5 & 40.3 & 34.2 & 42.3 & 36.4 & 58.2 & 66.7 & 45.2 & 47.0 & 50.7  \\
\mfrand & \xmark& 63.8 & 55.1 & 59.5 & 59.5 & 38.4 & 21.4 & 51.8 & 51.6 & 63.5 & 42.8 & 59.8 & 44.9 & 35.0 & 45.5 & 42.0 & 58.4 & 69.8 & 46.7 & 48.6 & 53.1  \\ 
\mfrandkmeans &\cmark & 60.4 & 55.5 & 61.5 & 59.1 & 32.3 & 10.5 & 43.5 & 47.6 & 60.0 & 25.5 & 29.7 & 44.2 & 21.0 & 34.9 & 36.6 & 46.5 & 68.3 & 40.1 & 46.3 & 47.5 \\
\mfrandkmeans &\xmark & 61.3 & 59.4 & 63.5 & 61.4 & 39.3 & 11.4 & 49.9 & 51.4 & 62.4 & 28.8 & 35.3 & 47.8 & 26.6 & 39.2 & 43.1 & 45.7 & 69.1 & 42.7 & 48.2 & 49.8 \\
\mfrandnn &\cmark &     62.9 & 56.4 & 63.2 & 60.8 & 24.5 & 65.0 & 49.0 & 46.1 & 58.1 & 40.7 & 56.9 & 49.0 & 29.5 & 46.6 & 35.0 & 35.0 & 65.5 & 15.5 & 43.2 & 38.8 \\
\mfrandnn &\xmark &     64.1 & 59.6 & 65.8 & 63.2 & 26.8 & 34.2 & 56.6 & 49.2 & 59.8 & 41.5 & 63.4 & 53.7 & 33.8 & 46.6 & 39.6 & 36.1 & 67.1 & 18.9 & 49.6 & 42.3 \\
\cdashline{1-21} \vspace{-3mm} \\
\textsc{FMT-LexInit}  &  & 	64.3 & 47.6 & 67.5 & 59.8 & 47.0 & 28.5 & 68.9 & 53.4 & 66.0 & 43.3 & 49.1 & 57.5 & 43.0 & 50.7 & 34.6 & 32.1 & 62.6 & 45.9 & 50.4 & 45.2 \\
\ellex & \cmark & 62.0 & 68.0 & 67.1 & 65.7 & 40.7 & 47.5 & 52.1 & 54.3 & 58.4 & 35.8 & 53.8 & 53.5 & 24.8 & 46.8 & 46.1 & 59.2 & 61.5 & 42.1 & 52.4 & 52.3 \\
\ellex & \xmark & 67.5 & 70.0 & 69.5 & 69.0 & 42.9 & 47.7 & 63.4 & 55.7 & 63.6 & 41.4 & 53.6 & 55.4 & 25.1 & 49.9 & 46.5 & 58.6 & 61.8 & 44.8 & 52.3 & 52.8 \\
\mflex &\cmark  & 63.3 & 63.4 & 61.7 & 62.8 & 41.0 & 58.9 & 53.6 & 50.1 & 64.7 & 40.3 & 54.8 & 53.9 & 31.3 & 49.8 & 33.5 & 38.9 & 67.8 & 42.4 & 47.3 & 46.0\\
\mflex & \xmark & 64.7 & 64.3 & 63.4 & 64.1 & 46.3 & 58.1 & 61.3 & 54.6 & 65.7 & 46.4 & 53.1 & 55.5 & 36.5 & 53.1 & 36.1 & 44.4 & 70.2 & 43.1 & 52.2 & 49.2\\ 
\mflexkmeans &\cmark & 64.8 & 61.9 & 64.2 & 63.7 & 35.9 & 52.0 & 56.0 & 51.7 & 57.5 & 41.5 & 50.3 & 51.1 & 30.3 & 47.4 & 46.2 & 52.6 & 69.2 & 44.1 & 49.2 & 52.3 \\
\mflexkmeans &\xmark & 66.5 & 62.9 & 65.4 & 64.9 & 43.3 & 62.0 & 57.5 & 55.0 & 60.7 & 45.2 & 50.1 & 55.9 & 35.1 & 51.6 & 50.5 & 52.4 & 70.9 & 45.5 & 51.6 & 54.2 \\
\mflexnn &\cmark &     63.4 & 66.5 & 63.4 & 64.4 & 44.7 & 38.6 & 40.7 & 52.0 & 58.7 & 42.5 & 52.9 & 54.6 & 34.8 & 46.6 & 46.4 & 53.4 & 64.3 & 44.4 & 42.5 & 50.2  \\
\mflexnn &\xmark &     64.7 & 68.3 & 63.7 & 65.5 & 47.9 & 54.9 & 52.0 & 54.6 & 59.8 & 45.3 & 57.0 & 57.3 & 38.7 & 51.9 & 48.9 & 51.9 & 68.6 & 46.2 & 45.3 & 52.2  \\
\bottomrule
\end{tabular}
} 

\caption{  Mean $F_1$ NER test results averaged over 5 runs transferring from high resource language  \textbf{Arabic} to the low resource languages. \textit{12Ad} indicates whether (\cmark) or not (\xmark) an adapter is placed in the 12th transformer layer.  The top group (first three rows) includes models which leverage the original tokenizer which is not specialized for the target language. The second group (last 18 rows) include models with new tokenizers. Here we separate models with randomly initialized embeddings ($*$-\textsc{RandInit}) from models with lexical initialization ($*$-\textsc{LexInit}) by the dashed line. We additionally present the results for full model fine-tuning (\textsc{FMT}-$*$).  
}
\label{tab:appendix_zeroshot_ner}

\end{table*}

\begin{table*}
\begin{subtable}[t]{\linewidth}
\centering
\footnotesize
\resizebox{\textwidth}{!}{
\begin{tabular}{lrrr|rrrr|r}
\toprule
    & \multicolumn{3}{c}{Seen Languages} & \multicolumn{4}{c}{Unseen Languages but Covered Scripts} & \multicolumn{1}{c}{New Script} \\ 
	 & 	 \multicolumn{1}{c}{hi} & \multicolumn{1}{c}{ur} & \multicolumn{1}{c}{\textbf{Avg}} & \multicolumn{1}{c}{bh} & \multicolumn{1}{c}{myv} &  \multicolumn{1}{c}{wo} & \multicolumn{1}{c}{\textbf{Avg}} & \multicolumn{1}{c}{am}   \\
		 & 	 \multicolumn{1}{c}{UAS / LAS} & \multicolumn{1}{c}{UAS / LAS} & \multicolumn{1}{c}{UAS / LAS} & \multicolumn{1}{c}{UAS / LAS} & \multicolumn{1}{c}{UAS / LAS} &  \multicolumn{1}{c}{UAS / LAS} & \multicolumn{1}{c}{UAS / LAS} & \multicolumn{1}{c}{UAS / LAS}   \\
\midrule
mBERT &  48.0 / 34.4 & 36.4 / 23.9 & 42.2 / 29.2 & 33.0 / 18.3 & 33.6 / 14.3 & 33.0 / \, 9.3 & 33.2 / 14.0 & 15.5 / 12.6\\

\textsc{MAD-X 2.0} & 44.6 / 31.2 & 37.6 / 23.9 & 41.1 / 27.5 & 29.5 / 16.1 & 60.5 / 43.2 & 50.5 / 33.8 & 46.8 / 31.1 & 11.7 / \, 9.6  \\
\midrule
\elrand	 &  45.9 / 32.5 & 38.2 / 23.9 & 42.1 / 28.2 & 29.9 / 13.6 & 57.8 / 35.0 & 33.2 / 11.1 & 40.3 / 19.9 & 26.4 / 13.0 \\
\mfrand &  45.3 / 33.3 & 37.8 / 25.2 & 41.6 / 29.3 & 26.6 / 13.3 & 62.7 / 41.5 & 42.2 / 27.2 & 43.8 / 27.3 & 32.7 / 15.5 \\

\mfrandkmeans & 46.3 / 33.1 & 37.8 / 25.6 & 42.1 / 29.3 & 31.4 / 16.0 & 62.9 / 42.8 & 24.9 / 13.7 & 39.7 / 24.1 & 29.7 / 14.9\\
\mfrandnn &     42.6 / 30.5 & 33.7 / 22.6 & 38.1 / 26.5 & 33.2 / 17.8 & 61.6 / 41.5 & 38.3 / 22.2 & 44.4 / 27.2 & 28.4 / 12.7\\

\cdashline{1-9} \vspace{-2.5mm} \\
\ellex & 47.1 / 33.1 & 38.1 / 24.9 & 42.6 / 29.0 & 31.3 / 17.2 & 63.3 / 40.9 & 51.3 / 34.0 & 48.6 / 30.7 & 32.4 / \, 9.1 \\
\mflex & 45.8 / 33.5 & 38.1 / 26.2 & 42.0 / 29.8 & 33.2 / 17.6 & 62.6 / 42.4 & 50.0 / 35.2 & 48.6 / 31.8 & 35.1 / 16.1 \\
\mflexkmeans & 46.0 / 33.4 & 38.8 / 26.5 & 42.4 / 30.0 & 33.1 / 17.7 & 64.2 / 44.5 & 52.6 / 35.9 & 50.0 / 32.7 & 27.7 / 14.1\\
\mflexnn &     45.1 / 32.9 & 36.8 / 25.1 & 41.0 / 29.0 & 32.1 / 17.0 & 62.3 / 42.9 & 45.2 / 29.6 & 46.5 / 29.8 & 29.6 / 15.3 \\
\bottomrule
\end{tabular}
}
\caption{  Zero-shot DP scores. Source language:  \textbf{English}. 
}
\label{tab:mainresults_UD_english}
\end{subtable}

\vspace{0.5em}

\begin{subtable}[t]{\linewidth}
\centering
\footnotesize
\resizebox{\textwidth}{!}{
\begin{tabular}{lrrr|rrrr|r}
\toprule
    & \multicolumn{3}{c}{Seen Languages} & \multicolumn{4}{c}{Unseen Languages but Covered Scripts} & \multicolumn{1}{c}{New Script} \\ 
	 & 	 \multicolumn{1}{c}{hi} & \multicolumn{1}{c}{ur} & \multicolumn{1}{c}{\textbf{Avg}} & \multicolumn{1}{c}{bh} & \multicolumn{1}{c}{myv} &  \multicolumn{1}{c}{wo} & \multicolumn{1}{c}{\textbf{Avg}} & \multicolumn{1}{c}{am}   \\
		 & 	 \multicolumn{1}{c}{UAS / LAS} & \multicolumn{1}{c}{UAS / LAS} & \multicolumn{1}{c}{UAS / LAS} & \multicolumn{1}{c}{UAS / LAS} & \multicolumn{1}{c}{UAS / LAS} &  \multicolumn{1}{c}{UAS / LAS} & \multicolumn{1}{c}{UAS / LAS} & \multicolumn{1}{c}{UAS / LAS}   \\
\midrule
mBERT &  53.1 / 28.7 & 38.7 / 19.4 & 45.9 / 24.1 & 37.8 / 16.8 & 33.0 / 16.1 & 28.2 / \, 9.2 & 33.0 / 14.0 & \, 7.7 / \, 0.6\\

\textsc{MAD-X 2.0} &  48.2 / 25.0 & 39.2 / 19.1 & 43.7 / 22.0 & 33.4 / 14.8 & 52.5 / 32.4 & 37.6 / 20.1 & 41.2 / 22.5 & 12.6 / \, 9.8 \\
\midrule
\elrand	 & 46.2 / 22.7 & 34.1 / 15.7 & 40.1 / 19.2 & 30.0 / 10.0 & 49.6 / 25.8 & 22.7 / \, 6.9 & 34.1 / 14.2 & 27.3 / 11.0  \\
\mfrand &  48.0 / 25.5 & 36.0 / 17.7 & 42.0 / 21.6 & 29.9 / 12.2 & 54.5 / 32.0 & 34.6 / 17.3 & 39.7 / 20.5 & 33.8 / 13.7 \\
\mfrandkmeans & 46.5 / 25.0 & 35.7 / 17.8 & 41.1 / 21.4 & 29.5 / 11.4 & 54.9 / 32.6 & 22.8 / 11.4 & 35.7 / 18.5 & 34.3 / 13.3 \\
\mfrandnn &     44.3 / 22.9 & 31.7 / 15.3 & 38.0 / 19.1 & 32.2 / 13.6 & 53.6 / 31.7 & 29.9 / 15.6 & 38.6 / 20.3 & 32.9 / 11.7 \\
\cdashline{1-9} \vspace{-2.5mm} \\
\ellex & 47.1 / 24.6 & 36.9 / 17.7 & 42.0 / 21.1 & 31.6 / 14.0 & 53.5 / 29.6 & 40.3 / 20.9 & 41.8 / 21.5 & 34.4 / 10.8 \\
\mflex &  47.4 / 26.1 & 36.8 / 18.3 & 42.1 / 22.2 & 31.3 / 13.9 & 54.3 / 32.7 & 39.5 / 21.2 & 41.7 / 22.6 & 37.6 / 14.8\\
\mflexkmeans & 47.6 / 25.8 & 38.4 / 18.9 & 43.0 / 22.4 & 33.5 / 14.3 & 54.5 / 32.7 & 40.5 / 22.5 & 42.8 / 23.2 & 28.4 / 11.2 \\
\mflexnn &     45.0 / 24.8 & 36.7 / 18.4 & 40.9 / 21.6 & 32.2 / 14.6 & 53.0 / 31.3 & 35.0 / 19.1 & 40.1 / 21.7 & 34.2 / 15.4  \\
\bottomrule
\end{tabular}
}
\caption{Zero-shot DP scores. Source language:  \textbf{Chinese}. 
}
\label{tab:mainresults_UD_chinese}
\end{subtable}

\vspace{0.5em}

\begin{subtable}[t]{\linewidth}
\centering
\footnotesize
\resizebox{\textwidth}{!}{
\begin{tabular}{lrrr|rrrr|r}
\toprule
    & \multicolumn{3}{c}{Seen Languages} & \multicolumn{4}{c}{Unseen Languages but Covered Scripts} & \multicolumn{1}{c}{New Script} \\ 
	 & 	 \multicolumn{1}{c}{hi} & \multicolumn{1}{c}{ur} & \multicolumn{1}{c}{\textbf{Avg}} & \multicolumn{1}{c}{bh} & \multicolumn{1}{c}{myv} &  \multicolumn{1}{c}{wo} & \multicolumn{1}{c}{\textbf{Avg}} & \multicolumn{1}{c}{am}   \\
		 & 	 \multicolumn{1}{c}{UAS / LAS} & \multicolumn{1}{c}{UAS / LAS} & \multicolumn{1}{c}{UAS / LAS} & \multicolumn{1}{c}{UAS / LAS} & \multicolumn{1}{c}{UAS / LAS} &  \multicolumn{1}{c}{UAS / LAS} & \multicolumn{1}{c}{UAS / LAS} & \multicolumn{1}{c}{UAS / LAS}   \\
\midrule
mBERT &  57.0 / 40.7 & 43.7 / 29.8 & 50.3 / 35.2 & 45.9 / 30.6 & 27.7 / \, 9.3 & 18.9 / \, 2.3 & 30.9 / 14.0 & 10.0 / \, 0.9\\

\textsc{MAD-X 2.0} & 55.5 / 38.3 & 45.4 / 30.4 & 50.5 / 34.3 & 42.6 / 27.2 & 37.9 / 19.7 & 22.2 / \, 6.1 & 34.2 / 17.7 & 12.3 / \, 9.7  \\
\midrule
\elrand	 &  55.7 / 38.4 & 44.5 / 28.1 & 50.1 / 33.2 & 42.4 / 24.9 & 31.8 / \, 9.5 & 16.3 / \, 1.9 & 30.2 / 12.1 & 36.9 / 14.3 \\
\mfrand &  53.1 / 36.7 & 42.7 / 27.4 & 47.9 / 32.1 & 38.5 / 23.9 & 36.6 / 14.6 & 21.1 / \, 4.3 & 32.0 / 14.3 & 36.6 / 15.2 \\
\mfrandkmeans & 54.6 / 38.0 & 44.6 / 28.7 & 49.6 / 33.3 & 41.3 / 25.9 & 36.6 / 15.5 & 14.1 /  \, 2.3 & 30.7 / 14.6 & 33.1 / 14.0 \\
\mfrandnn &     52.2 / 35.8 & 40.9 / 26.7 & 46.6 / 31.3 & 42.9 / 27.5 & 35.5 / 15.0 & 18.8 /  \, 4.3 & 32.4 / 15.6 & 29.0 / 10.2 \\
\cdashline{1-9} \vspace{-2.5mm} \\
\ellex & 52.9 / 36.3 & 45.1 / 29.1 & 49.0 / 32.7 & 40.6 / 27.1 & 34.9 / 13.3 & 22.6 / \, 5.3 & 32.7 / 15.2 & 37.6 / 11.6 \\
\mflex & 53.1 / 36.8 & 44.2 / 28.6 & 48.7 / 32.7 & 41.5 / 27.0 & 36.1 / 14.8 & 23.9 / \, 5.6 & 33.8 / 15.8 & 31.2 / 12.5 \\
\mflexkmeans & 53.4 / 37.8 & 45.7 / 30.4 & 49.6 / 34.1 & 41.9 / 27.0 & 36.9 / 16.2 & 23.6 /  \, 5.6 & 34.1 / 16.3 & 40.9 / 16.5 \\
\mflexnn &     53.5 / 37.9 & 44.5 / 29.2 & 49.0 / 33.5 & 41.3 / 27.0 & 35.6 / 16.7 & 20.4 /  \, 4.8 & 32.4 / 16.1 & 32.6 / 14.5  \\
\bottomrule
\end{tabular}
}
\caption{Zero-shot DP scores. Source language:  \textbf{Japanese}. 
}
\label{tab:mainresults_UD_japanese}
\end{subtable}

\vspace{0.5em}
 
 \begin{subtable}[t]{\linewidth}
\centering
\footnotesize
\resizebox{\textwidth}{!}{
\begin{tabular}{lrrr|rrrr|r}
\toprule
    & \multicolumn{3}{c}{Seen Languages} & \multicolumn{4}{c}{Unseen Languages but Covered Scripts} & \multicolumn{1}{c}{New Script} \\ 
	 & 	 \multicolumn{1}{c}{hi} & \multicolumn{1}{c}{ur} & \multicolumn{1}{c}{\textbf{Avg}} & \multicolumn{1}{c}{bh} & \multicolumn{1}{c}{myv} &  \multicolumn{1}{c}{wo} & \multicolumn{1}{c}{\textbf{Avg}} & \multicolumn{1}{c}{am}   \\
		 & 	 \multicolumn{1}{c}{UAS / LAS} & \multicolumn{1}{c}{UAS / LAS} & \multicolumn{1}{c}{UAS / LAS} & \multicolumn{1}{c}{UAS / LAS} & \multicolumn{1}{c}{UAS / LAS} &  \multicolumn{1}{c}{UAS / LAS} & \multicolumn{1}{c}{UAS / LAS} & \multicolumn{1}{c}{UAS / LAS}   \\
\midrule
mBERT & 25.1 / 14.2 & 20.9 / \, 8.9 & 23.0 / 11.6 & 18.9 / \, 8.7 & 24.5 / 12.3 & 29.9 / \, 9.1 & 24.4 / 10.0 & \, 7.3 / \, 1.1 \\

\textsc{MAD-X 2.0} &  19.8 /  \,  9.8 & 17.9 / \, 7.0 & 18.8 / \, 8.4 & 18.8 /\,6.7 & 39.7 / 23.7 & 31.2 / 13.6 & 29.9 / 14.6 & 19.4 / \, 1.5 \\
\midrule
\elrand	 &  19.7 / 10.0 & 15.0 / \, 6.0 & 17.4 / \, 8.0 & 17.5 / \, 6.6 & 42.2 / 25.5 & 25.2 / \, 8.9 & 28.3 / 13.7 & 30.0 / \, 6.7 \\
\mfrand &  20.3 / \, 9.4 & 18.4 / \, 7.3 & 19.4 / \, 8.4 & 19.7 / \, 5.3 & 36.7 / 20.7 & 33.1 / 13.3 & 29.8 / 13.1 & 34.0 / \, 8.9 \\
\mfrandkmeans & 25.1 / 11.9 & 20.9 /  \, 7.9 & 23.0 / 9.9 & 21.0 /  \, 5.5 & 40.0 / 23.8 & 25.1 / 10.0 & 28.7 / 13.1 & 32.7 /  \, 7.6 \\
\mfrandnn   &   23.8 / 11.4 & 15.3 /  \, 6.3 & 19.5 /  \, 8.9 & 19.1 /  \, 6.1 & 40.3 / 22.6 & 31.2 / 12.7 & 30.2 / 13.8 &  \, 0.0 /  \, 5.8 \\
\cdashline{1-9} \vspace{-2.5mm} \\
\ellex & 20.5 / 10.6 & 15.8 / \, 6.8 & 18.1 / \, 8.7 & 19.0 / \, 6.7 & 43.3 / 26.9 & 37.2 / 16.6 & 33.2 / 16.8 & 31.4 / \, 8.3 \\
\mflex & 22.0 / 10.5 & 16.9 / \, 7.1 & 19.5 / \, 8.8 & 18.9 / \, 7.3 & 42.8 / 25.6 & 36.9 / 16.1 & 32.9 / 16.3 & 35.4 / \, 8.0 \\
\mflexkmeans    & 25.2 / 12.8 & 21.7 /  \, 8.8 & 23.4 / 10.8 & 15.1 /  \, 5.6 & 31.6 / 17.0 & 38.0 / 17.5 & 28.2 / 13.4 & 28.7 /  \, 6.6 \\
\mflexnn        & 25.3 / 12.5 & 21.0 /  \, 8.3 & 23.2 / 10.4 & 18.9 /  \, 6.8 & 39.9 / 23.5 & 34.5 / 15.9 & 31.1 / 15.4 & 32.4 /  \, 7.7 \\
\bottomrule
\end{tabular}
}
\caption{Zero-shot DP scores. Source language:  \textbf{Arabic}. 
}
\label{tab:mainresults_UD_arabic}
\end{subtable}
\caption{ 
Mean UAS and LAS test results for UD averaged over 5 runs transferring from the high-resource source languages (a) \textbf{English}, (b) \textbf{Chinese}, (c) \textbf{Japanses}, and (d) \textbf{Arabic}.  The top group (first two rows) includes models which leverage the original tokenizer which is not specialized for the target language. The second group (last four rows) include models with new tokenizers. Here we separate models with randomly initialized embeddings ($*$-\textsc{RandInit}) from models with lexical initialization ($*$-\textsc{LexInit}) by the dashed line. 
}
\label{tab:mainresults_UD_appendix}
\end{table*}

\begin{table*}[]
\centering
\resizebox{\textwidth}{!}{%
\begin{tabular}{llr}
\toprule
Language & Language code & Script \\
\midrule
Afrikaans & af & Latin \\
Albanian & sq & Latin \\
Arabic & ar & Arabic \\
Aragonese & an & Latin \\
Armenian & hy & Armenian \\
Asturian & ast & Latin \\
Azerbaijani & az & Latin \\
Bashkir & ba & Cyrillic \\
Basque & eu & Latin \\
Bavarian & bar & Latin \\
Belarusian & be & Cyrillic \\
Bengali & bn & Bengali \\
Bishnupriya-manipuri & bpy & Bengali \\
Bosnian & bs & Latin \\
Breton & br & Latin \\
Bulgarian & bg & Cyrillic \\
Burmese & my & Myanmar \\
Catalan & ca & Latin \\
Cebuano & ceb & Latin \\
Chechen & ce & Cyrillic \\
Chinese-simplified & zh-Hans & Chinese \\
Chinese-traditional & zh-Hant & Chinese \\
Chuvash & cv & Cyrillic \\
Croatian & hr & Latin \\
Czech & cs & Latin \\
Danish & da & Latin \\
Dutch & nl & Latin \\
English & en & Latin \\
Estonian & et & Latin \\
Finnish & fi & Latin \\
French & fr & Latin \\
Galician & gl & Latin \\
Georgian & ka & Georgian \\
German & de & Latin \\
Greek & el & Greek \\
Gujarati & gu & Gujarati \\
Haitian & ht & Latin \\
Hebrew & he & Hebrew \\
Hindi & hi & Devanagari \\
Hungarian & hu & Latin \\
Icelandic & is & Latin \\
Ido & io & Latin \\
Indonesian & id & Latin \\
Irish & ga & Latin \\
Italian & it & Latin \\
Japanese & ja & Japanese \\
Javanese & jv & Latin \\
Kannada & kn & Kannada \\
Kazakh & kk & Cyrillic \\
Kirghiz & ky & Cyrillic \\
Korean & ko & Korean \\
Latin & la & Latin \\
\bottomrule
\end{tabular}

\quad

\begin{tabular}{llr}
\toprule
Language & Language code & Script \\
\midrule
Latvian & lv & Latin \\
Lithuanian & lt & Latin \\
Lombard & lmo & Latin \\
Low-saxon & nds & Latin \\
Luxembourgish & lb & Latin \\
Macedonian & mk & Cyrillic \\
Malagasy & mg & Latin \\
Malay & ms & Latin \\
Malayalam & ml & Malayalam \\
Marathi & mr & Devanagari \\
Minangkabau & min & Latin \\
Mongolian & mn & Cyrillic \\
Nepali & ne & Devanagari \\
Newar & new & Devanagari \\
Norwegian-bokmal & nb & Latin \\
Norwegian-nynorsk & nn & Latin \\
Occitan & oc & Latin \\
Persian & fa & Arabic \\
Piedmontese & pms & Latin \\
Polish & pl & Latin \\
Portuguese & pt & Latin \\
Punjabi & pa & Gurmukhi \\
Romanian & ro & Latin \\
Russian & ru & Cyrillic \\
Scots & sco & Latin \\
Serbian & sr & Cyrillic \\
Serbo-croatian & hbs & Latin \\
Sicilian & scn & Latin \\
Slovak & sk & Latin \\
Slovenian & sl & Latin \\
South-azerbaijani & azb & Arabic \\
Spanish & es & Latin \\
Sundanese & su & Latin \\
Swahili & sw & Latin \\
Swedish & sv & Latin \\
Tagalog & tl & Latin \\
Tajik & tg & Cyrillic \\
Tamil & ta & Tamil \\
Tatar & tt & Cyrillic \\
Telugu & te & Telugu \\
Thai & th & Thai \\
Turkish & tr & Latin \\
Ukrainian & uk & Cyrillic \\
Urdu & ur & Arabic \\
Uzbek & uz & Latin \\
Vietnamese & vi & Latin \\
Volapuk & vo & Latin \\
Waray-waray & war & Latin \\
Welsh & cy & Latin \\
West & fy & Latin \\
Western-punjabi & lah & Arabic \\
Yoruba & yo & Latin \\
\bottomrule
\end{tabular}
}
\caption{List of languages used in the pretraining. Taken from \citet{ChungGTR20}}
\label{table:list_of_languages}
\end{table*}

\end{document}